\def\year{2021}\relax
\documentclass[letterpaper]{article} 
\usepackage{aaai21}  
\usepackage{times}  
\usepackage{helvet} 
\usepackage{courier}  
\usepackage[hyphens]{url}  
\usepackage{graphicx} 
\usepackage{natbib}  
\usepackage{caption} 
\usepackage{booktabs}
\usepackage{amsmath}
\usepackage{amsfonts}
\usepackage{amssymb}
\usepackage{multirow}

\newtheorem{theorem}{Theorem}
\newtheorem{remark}{Remark}

\newcommand{\mbf}[1]{\mathbf{#1}}
\newcommand{\mcl}[1]{\mathcal{#1}}
\newcommand{\mbb}[1]{\mathbb{#1}}
\newcommand{\argmax}{\arg\!\max}
\newcommand{\argmin}{\arg\!\min}

\newcommand{\citesupp}[1]{Appendix~\ref{#1}}

\title{An Information-Theoretic Framework for Unifying Active Learning Problems}
\author {
        Quoc Phong Nguyen,\textsuperscript{\rm 1}
        Bryan Kian Hsiang Low,\textsuperscript{\rm 1}
        Patrick Jaillet\textsuperscript{\rm 2} \\
}
\affiliations {
    \textsuperscript{\rm 1}Dept. of Computer Science, National University of Singapore, Republic of Singapore\\
    \textsuperscript{\rm 2}Dept. of Electrical Engineering and Computer Science, MIT, USA\\
    \{qphong, lowkh\}@comp.nus.edu.sg, jaillet@mit.edu
}

\begin{document}

\maketitle

\begin{abstract}
This paper presents an information-theoretic framework for unifying active learning problems: \emph{level set estimation} (LSE), \emph{Bayesian optimization} (BO), and their generalized variant. We first introduce a novel active learning criterion that subsumes an existing LSE algorithm and achieves state-of-the-art performance in LSE problems with a continuous input domain. Then, by exploiting the relationship between LSE and BO, we design a competitive information-theoretic acquisition function for BO that has interesting connections to 
upper confidence bound and \emph{max-value entropy search} (MES). The latter connection reveals a drawback of MES which has important implications on not only MES but also on other MES-based acquisition functions. Finally, our unifying information-theoretic framework can be applied to solve a generalized problem of LSE and BO involving multiple level sets in a data-efficient manner. We empirically evaluate the performance of our proposed algorithms using synthetic benchmark functions, a real-world dataset, and in hyperparameter tuning of machine learning models.
\end{abstract}

\section{Introduction}
\label{sec:introduction}
\emph{Level set estimation} (LSE) is about determining a level set of an unknown function or, alternatively, a \emph{superlevel set} of the function (i.e., a region of inputs where the function values are of at least a known threshold) 
given a finite budget of expensive (possibly noisy) function evaluations~\cite{gotovos2013active}. It has important applications in environmental sensing/monitoring  where the unknown function is a (spatial) field of some quantity of interest (e.g., pH, temperature, and solar radiation)~\cite{galland2004synthetic}.
On the other hand, \emph{Bayesian optimization} (BO) has gained significant recognition in science and engineering fields~\citep{brochu10tut,calandra14,krause11,shahriari15,snoek12} for its effectiveness in optimizing a black-box objective function (i.e., without a closed-form expression/derivative) using a finite budget of expensive (possibly noisy) function evaluations.
At first glance, one may straightforwardly regard BO as LSE by setting the threshold as the maximum value of the objective function, i.e., the superlevel set is reduced to a set of maximizers. However, the \emph{unknown} maximum value in BO does not satisfy the  requirement of a \emph{known} threshold in LSE. 
This poses the  challenge of whether it is possible to develop a framework to unify  LSE and BO.
Though the work of~\citet{bogunovic16} has developed such a unified approach called truncated variance reduction, it is demonstrated  mainly on problems with a discrete input domain and requires enumerating over all inputs in a set of ``unclassified points'', which can be
prohibitively large in practice (or infinite when the input domain is continuous and not discretized). 
In contrast, our work here proposes the \emph{first  information-theoretic} framework for unifying both LSE and BO that can empirically outperform the state-of-the-art LSE criteria
and scale to real-world problems with a continuous input domain.

To shed light on the connection between LSE and BO, we propose to view BO as an active learning problem that involves actively estimating the superlevel set of the objective function with respect to an estimate of its maximum value; 
such a problem reduces to LSE when the maximum value is known instead. 
Improving the estimation of the superlevel set in turn refines the estimate of the maximum value. 
As the estimate approaches the true maximum value, the superlevel set becomes a set of the maximizers of the objective function.
Unfortunately, existing LSE criteria cannot be directly applied to BO since they either impose a noiseless assumption \cite{low12} or cannot handle an unknown threshold \cite{bryan2006active}.

A key contribution of our work here therefore lies in introducing a novel information-theoretic active learning criterion for LSE (Sec.~\ref{sec:beslse}) that can be exploited for designing a new acquisition function for BO with interesting connections to \emph{upper confidence bound} (UCB) and \emph{max-value entropy search} (MES) (Sec.~\ref{sec:besbo}). 
The latter connection reveals a drawback of MES  (Remark~\ref{rmk:besmpmestruncnorm}), which has important implications on not only MES, but also on other MES-based acquisition functions such as those handling multiple objectives \cite{syrine19,suzuki2019multi} or fidelities \cite{takeno2019multi}.

The other main contribution of our work is to show how our proposed unifying information-theoretic  framework can be applied to solve a generalized problem of LSE and BO involving multiple level sets/thresholds in a data-efficient manner. 
This problem, namely \emph{implicit LSE} (Sec.~\ref{sec:ilse}), is about identifying a region of inputs whose function values differ from the (unknown) maximum value by at most a specified \emph{tolerance}. It is motivated from the estimation of \emph{hotspots} in environmental fields, which correspond to regions with large field measurements \cite{gotovos2013active}.
In summary, the specific contributions of our work include:
\begin{itemize}
\item A novel information-theoretic active learning criterion for LSE problems with a continuous input domain (Sec.~\ref{sec:beslse}), which subsumes an existing LSE criterion \cite{low12}
and empirically outperforms state-of-the-art LSE criteria \cite{bryan2006active,low12}
on synthetic benchmark functions and a real-world dataset (Sec.~\ref{subsec:experimentlse});
\item A new information-theoretic acquisition function for BO problems with interesting connections to UCB and MES; 
the latter connection reveals a drawback of using MES (Sec.~\ref{sec:besbo}).
We empirically evaluate the performance of our proposed BO algorithm using several synthetic benchmark functions, a real-world dataset, and in hyperparameter tuning of a logistic regression model and a convolutional neural network for image classification with MNIST and CIFAR-$10$ datasets 
(Sec.~\ref{subsec:experimentbo});
\item Applying our unifying information-theoretic  framework to solve the implicit LSE problem in a data-efficient manner 
(Sec.~\ref{sec:ilse}).
\end{itemize}
\section{Gaussian Process (GP)}
\label{sec:gp}
Let the unknown objective function be denoted as $f: \mcl{X} \rightarrow \mbb{R}$ over a bounded input domain $\mcl{X} \subset \mbb{R}^d$.
An LSE/BO algorithm repeatedly selects an input query $\mbf{x} \in \mcl{X}$ for evaluating $f$ to obtain a noisy observation $y_{\mbf{x}} \triangleq f(\mbf{x}) + \epsilon_{\mbf{x}}$ of its function value $f(\mbf{x})$ corrupted by an additive Gaussian noise $\epsilon_{\mbf{x}} \sim \mcl{N}(0,\sigma_n^2)$ with noise variance $\sigma_n^2$. 
Since it is expensive to evaluate $f$, the goal of LSE (BO) is to strategically select input queries for finding the level/superlevel set (global maximizer(s)) as rapidly as possible.
To achieve this, we model $f$ using a GP: Let $\{ f(\mbf{x}') \}_{\mbf{x}' \in \mcl{X}}$ denote a GP, i.e., every finite subset of $\{ f(\mbf{x}') \}_{\mbf{x}' \in \mcl{X}}$ follows a multivariate Gaussian distribution \cite{rasmussen06}. Then, the GP is fully specified by its \emph{prior} mean $\mbb{E}[f(\mbf{x}')]$ and covariance $k_{\mbf{x}'\mbf{x}''} \triangleq \text{cov}[f(\mbf{x}'), f(\mbf{x}'')]$ for all $\mbf{x}', \mbf{x}'' \in \mcl{X}$; the latter can be defined by, for example, the widely-used squared exponential kernel
$k_{\mbf{x}'\mbf{x}''} \triangleq \sigma_s^2  \exp(-0.5(\mbf{x}' - \mbf{x}'')^\top{\Lambda}^{-2}(\mbf{x}' - \mbf{x}''))$
where ${\Lambda} \triangleq \mathrm{diag}(\ell_1, \ldots, \ell_d)$ and $\sigma_s^2$ are its length-scale and signal variance hyperparameters, respectively. For notational simplicity (and w.l.o.g.), the prior mean is assumed to be zero. 
Given a column vector $\mbf{y}_{\mathcal{D}}\triangleq (y_{\mbf{x}'})^{\top}_{\mbf{x}'\in\mcl{D}}$ of noisy observations from evaluating $f$ at a set $\mcl{D}$ of input queries selected in previous iterations, the GP posterior belief of the function value at any input query $\mbf{x}$ is a Gaussian $p(f(\mbf{x}) |\mbf{y}_{\mathcal{D}}) = \mcl{N}(f(\mbf{x})| \mu_{\mbf{x}}, \sigma_{\mbf{x}}^2)$ with the following \emph{posterior} mean $\mu_{\mbf{x}}$ and variance $\sigma_{\mbf{x}}^2$:
\begin{equation}
\begin{array}{rl}
\displaystyle \mu_{\mbf{x}} 
	\triangleq \hspace{-2.4mm}&
\mbf{K}_{\mbf{x}\mcl{D}}(\mbf{K}_{\mcl{D}\mcl{D}}+\sigma^2_n \mbf{I})^{-1}
	\mbf{y}_{\mcl{D}}\vspace{1mm}\\
\displaystyle \sigma_{\mbf{x}}^2 
	\triangleq \hspace{-2.4mm}& k_{\mbf{x} \mbf{x}}
	- \mbf{K}_{\mbf{x}\mcl{D}}(\mbf{K}_{\mcl{D}\mcl{D}}+\sigma^2_n \mbf{I})^{-1}
	\mbf{K}_{\mcl{D}\mbf{x}}
\end{array}
\label{eq:gppost}
\end{equation}
where 
$\mbf{K}_{\mbf{x}\mcl{D}}\triangleq(k_{\mbf{x}\mbf{x}'})_{\mbf{x}'\in \mcl{D}}$, $\mbf{K}_{\mcl{D}\mcl{D}}\triangleq(k_{\mbf{x}'\mbf{x}''})_{\mbf{x}', \mbf{x}''\in \mcl{D}}$, $\mbf{K}_{\mcl{D}\mbf{x}}\triangleq \mbf{K}^{\top}_{\mbf{x}\mcl{D}}$, and $\mbf{I}$ is an identity matrix.
Then, $p(y_{\mbf{x}} |\mbf{y}_{\mathcal{D}}) = \mcl{N}(y_{\mbf{x}}| \mu_{\mbf{x}}, \sigma_{+}^2 \triangleq \sigma_{\mbf{x}}^2+\sigma^2_n)$.
\section{Binary Entropy Search (BES) for\\
Level Set Estimation (LSE)}
\label{sec:beslse}
An LSE algorithm repeatedly selects the next input query $\mbf{x} \in \mcl{X}$ for evaluating $f$ to maximize some active learning criterion based on the GP posterior belief of $f$ given the observations $\mbf{y}_{\mcl{D}}$ obtained in previous iterations such that the superlevel set $\mcl{X}_{f_\circ}^+ \triangleq \{\mbf{x}' \in \mcl{X}| f(\mbf{x}') \ge f_\circ\}$ of $f$ w.r.t.~a given threshold $f_\circ$ can be found as rapidly as possible.

In this section, we  propose an information-theoretic active learning criterion for LSE which measures the information gain on the superlevel set $\mcl{X}_{f_\circ}^+$ from evaluating $f$ at input query $\mbf{x}$ to observe $y_{\mbf{x}}$. 
Let $\gamma_{\mbf{x}}^\circ$ denote an indicator variable of label $-1$ if $\mbf{x} \in \mcl{X}_{f_\circ}^+$ (i.e., superlevel set), and label $1$ otherwise (i.e., $\mbf{x}$ is in sublevel set $\mcl{X}_{f_\circ}^- \triangleq \{\mbf{x}' \in \mcl{X}| f(\mbf{x}') < f_\circ\}$).
We can view $\gamma_{\mbf{x}}^\circ$ as a class label of $\mbf{x}$ and LSE as a binary classification problem that classifies whether each $\mbf{x}\in \mcl{X}$ 
is in the superlevel set
$\mcl{X}_{f_\circ}^+$ or the sublevel set $\mcl{X}_{f_\circ}^-$.
Let $\gamma_{\mcl{X}}^\circ \triangleq (\gamma_{\mbf{x}'}^\circ)_{\mbf{x}' \in \mcl{X}}$. 
The active learning criterion can therefore be measured as the mutual information $I(y_{\mbf{x}}; \gamma_{\mcl{X}}^\circ|\mbf{y}_{\mcl{D}}, f_\circ)$
which cannot be evaluated tractably with a continuous $\mcl{X}$. 
So, we simplify it to the information gain on class label $\gamma_{\mbf{x}}^\circ$ from evaluating $f$ at input query $\mbf{x}$ to observe $y_{\mbf{x}}$:
\begin{equation}
\hspace{-1.7mm}
\begin{array}{l}
\displaystyle \alpha_{\text{BES}}(\mbf{x}, \mbf{y}_{\mcl{D}}) \displaystyle\triangleq I(y_{\mbf{x}}; \gamma_{\mbf{x}}^\circ|\mbf{y}_{\mcl{D}}, f_\circ)
	\vspace{1mm}\\ 
	\displaystyle = H( p(\gamma_{\mbf{x}}^\circ|\mbf{y}_{\mcl{D}}, f_\circ) )\hspace{-0.7mm} -\hspace{-0.7mm} \mbb{E}_{ p(y_{\mbf{x}}|\mbf{y}_{\mcl{D}}, f_\circ)}\hspace{-0.8mm}\left[ H( p(\gamma_{\mbf{x}}^\circ|\mbf{y}_{\mcl{D} \cup \{\mbf{x}\}}, f_\circ) ) \right]
\end{array}
\label{eq:bes}
\end{equation}
where the \emph{prior entropy} of $\gamma_{\mbf{x}}^\circ$ is defined as
$$
H( p(\gamma_{\mbf{x}}^\circ|\mbf{y}_{\mcl{D}}, f_\circ) ) \triangleq - \sum_{\gamma_{\mbf{x}}^\circ} p(\gamma_{\mbf{x}}^\circ|\mbf{y}_{\mcl{D}}, f_\circ) \log p(\gamma_{\mbf{x}}^\circ|\mbf{y}_{\mcl{D}}, f_\circ)
$$ 
and the \emph{posterior entropy} $H( p(\gamma_{\mbf{x}}^\circ|\mbf{y}_{\mcl{D} \cup \{\mbf{x}\}}, f_\circ))$ of $\gamma_{\mbf{x}}^\circ$ given $y_{\mbf{x}}$ is defined in a similar manner.
Since $\gamma_{\mbf{x}}^\circ$ is binary, our active learning criterion $\alpha_{\text{BES}}(\mbf{x}, \mbf{y}_{\mcl{D}})$ is named \emph{binary entropy search} (BES).
Note that BES~\eqref{eq:bes} can be interpreted as the expected reduction in the uncertainty (entropy) of $\gamma_{\mbf{x}}^\circ$
from evaluating $f$ at input query $\mbf{x}$ to observe $y_{\mbf{x}}$. 
Though replacing $\gamma_{\mcl{X}}^\circ$ with $\gamma_{\mbf{x}}^\circ$ appears to be a simplification, BES demonstrates  state-of-the-art performance in our experiments (Sec.~\ref{subsec:experimentlse}).
Such a simplication is also commonly adopted by existing  acquisition functions for BO (e.g., \cite{suzuki2019multi,wang17mes}).
BES~\eqref{eq:bes} can be evaluated as follows: 
\begin{equation}
\hspace{-1.7mm}
\begin{array}{l}
\displaystyle \alpha_{\text{BES}}(\mbf{x}, \mbf{y}_{\mcl{D}}) \triangleq I(y_{\mbf{x}}; \gamma_{\mbf{x}}^\circ | \mbf{y}_{\mcl{D}}, f_{\circ})\\
= 
	\mbb{E}_{p(y_{\mbf{x}}| \mbf{y}_{\mcl{D}})}\hspace{-1mm} 
 \left[\hspace{-0.5mm} \displaystyle\sum_{\gamma_{\mbf{x}}^\circ} \Psi( \gamma_{\mbf{x}}^\circ\  g_{\mbf{x}}(y_{\mbf{x}}, f_{\circ})) 
 \log\hspace{-0.2mm} \frac{ \Psi( \gamma_{\mbf{x}}^\circ\  g_{\mbf{x}}(y_{\mbf{x}}, f_{\circ})) } 
		{\Psi(\gamma_{\mbf{x}}^\circ\  h_{\mbf{x}}(f_{\circ}))}
\hspace{-0.5mm}\right]
\end{array}
\label{eq:bescdf}
\end{equation}
where $g_{\mbf{x}}(y_{\mbf{x}}, f_{\circ}) \triangleq (\sigma_+^2 f_{\circ} - \sigma_n^2 \mu_{\mbf{x}} - \sigma_{\mbf{x}}^2 y_{\mbf{x}}) / (\sigma_{\mbf{x}}\sigma_n \sigma_+)$, $h_{\mbf{x}}(f_{\circ}) \triangleq (f_{\circ} - \mu_{\mbf{x}}) / \sigma_{\mbf{x}}$, and $\Psi$ denotes the c.d.f.~of the standard Gaussian distribution. Its derivation is shown in \citesupp{app:evallse}.
BES~\eqref{eq:bescdf} can thus be optimized w.r.t.~input query $\mbf{x}$ via stochastic gradient ascent.
\nocite{kingma15adam}
\nocite{kingma2013auto}

\begin{figure}
\begin{tabular}{@{}c|@{}c@{}}
\begin{tabular}{@{}c@{}}
\includegraphics[height=0.137\textwidth]{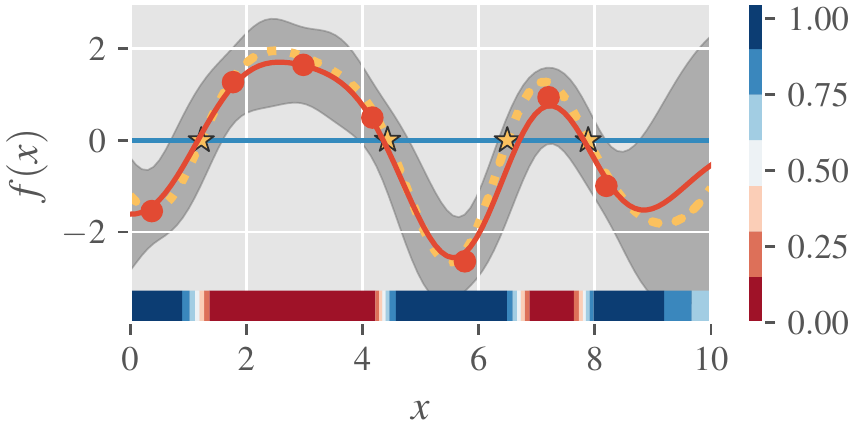}
\\
(a) GP posterior belief (gray)\vspace{1mm}
\\
\includegraphics[height=0.137\textwidth]{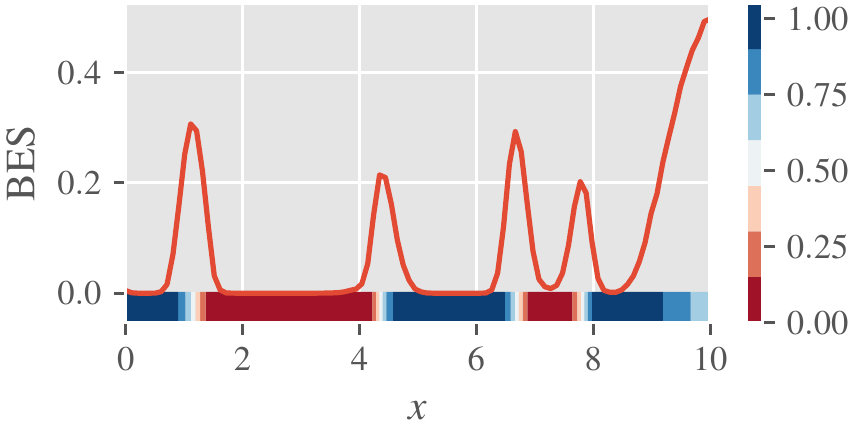}\\
(b) Values of BES
\end{tabular}
&
\begin{tabular}{c}
\includegraphics[height=0.25\textwidth]{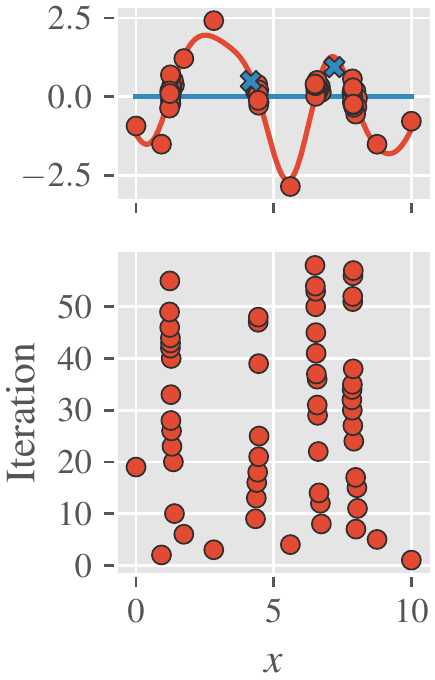}
\\
(c) Input queries
\end{tabular}
\end{tabular}
\caption{LSE with the threshold $f_\circ = 0$ as a binary classification problem that classifies if each $\mbf{x} \in [0,10]$ is in $\mcl{X}_{f_\circ}^+$ or $\mcl{X}_{f_\circ}^-$; $p(\mbf{x} \in \mcl{X}_{f_\circ}^- | \mbf{y}_{\mcl{D}})$ is shown on  $x$-axes of left plots. 
(a) The objective function $f$, level set, $7$ observations, GP posterior mean~\eqref{eq:gppost}, and  $f_\circ=0$ are plotted as a yellow dotted line, yellow stars, red circles, a red curve, and a blue line, respectively.
(b) Plot of values of BES based on GP posterior belief in Fig.~\ref{fig:class}a.
The bottom plot in (c): input queries (red circles) selected by BES w.r.t.~iteration no.; the top plot in (c): $f_\circ =0$, $2$ prior observations, and observations plotted as a blue line, blue crosses, and red circles, respectively.}
\label{fig:class}
\end{figure}

Fig.~\ref{fig:class}a shows LSE with the threshold $f_\circ = 0$ being viewed as a binary classification problem that classifies whether each $\mbf{x} \in [0,10]$ is in $\mcl{X}_{f_\circ}^+$ or $\mcl{X}_{f_\circ}^-$ and the level set w.r.t.~$f_\circ = 0$ is likely to be found on the decision boundary (i.e., white regions on the $x$-axis). 
Fig.~\ref{fig:class}b shows large values of BES
on the decision boundary that is likely to contain the level set, which is desirable.
Fig.~\ref{fig:class}c shows BES using about $10$ observations to explore and find roughly the level set w.r.t.~$f_\circ = 0$. Then, BES exploits by distributing its observations on the decision boundary (i.e., level set).
\begin{remark}[Special case of BES]
\emph{When the observation is noiseless (i.e., $y_{\mbf{x}} = f(\mbf{x})$ or $\sigma_n^2 = 0$), $\gamma_{\mbf{x}}^\circ$ is fully determined by the values of $f_\circ$ and $y_{\mbf{x}} = f(\mbf{x})$. Then, $p(\gamma_{\mbf{x}}^\circ| \mbf{y}_{\mcl{D} \cup\{\mbf{x}\}}, f_\circ)$ is either $0$ or $1$ and the posterior entropy term in \eqref{eq:bes} thus becomes $0$. So, BES reduces to the prior entropy term in \eqref{eq:bes} 
and the resulting active learning algorithm:
$\max_{\mbf{x}\in\mcl{X}}
H(p(\gamma_{\mbf{x}}^\circ |\mbf{y}_{\mcl{D}}, f_{\circ}))$ is called \emph{entropy maximization} (EM), as proposed by \citet{low12}.
EM can therefore be viewed as a special case of BES due to its noiseless observations. In other words, BES subsumes EM.}
\label{rmk:besaslow}
\end{remark}
\section{BES for Maximum Value Prediction (BES-MP) in Bayesian Optimization (BO)}
\label{sec:besbo}
A BO algorithm repeatedly selects the next input query $\mbf{x} \in \mcl{X}$ for evaluating $f$ to maximize some acquisition function based on the GP posterior belief of $f$ given the observations $\mbf{y}_{\mcl{D}}$ obtained in previous iterations such that the maximizer(s) of $f$ can be found as rapidly as possible.

Given an estimate $f_\star$ of the maximum value of $f$, the superlevel set $\mcl{X}_{f_\star}^+ \triangleq \{\mbf{x} \in \mcl{X}| f(\mbf{x}) \ge f_\star\}$ w.r.t.~the threshold $f_\star$ can be regarded as a set of potential maximizers. So, BO can be viewed as an active learning problem that involves actively estimating the superlevel set $\mcl{X}_{f_\star}^+$, which corresponds to an LSE problem. Therefore, we exploit our proposed BES criterion for LSE (Sec.~\ref{sec:beslse}) to design a new acquisition function for BO, specifically, the information gain on class label $\gamma_{\mbf{x}}^\star$ (i.e., indicator variable of label $-1$ if $\mbf{x} \in \mcl{X}_{f_\star}^+$, and label $1$ otherwise) 
from evaluating $f$ at input query $\mbf{x}$ to observe $y_{\mbf{x}}$:
$I(y_{\mbf{x}}; \gamma_{\mbf{x}}^\star| \mbf{y}_{\mcl{D}}, f_\star)$ which can be optimized via stochastic gradient ascent by replacing $\gamma_{\mbf{x}}^\circ$ and $f_\circ$ with $\gamma_{\mbf{x}}^\star$ and  $f_\star$ in \eqref{eq:bescdf}, respectively.
However, since the maximum value of $f$ is unknown, we estimate it with a set $\mcl{F}_\star$ of samples of the maximum value of functions drawn from the GP posterior belief~\eqref{eq:gppost}. 
These functions are drawn by applying the random Fourier feature approximation to GP \cite{rahimi08random}, which is widely used in existing information-theoretic acquisition functions \cite{pes,hoffman15opes,wang17mes}.
Then, we propose the acquisition function called \emph{BES for maximum value prediction} (BES-MP) by averaging our BES  criterion (for LSE) over the set $\mcl{F}_\star$ of maximum value samples:
\begin{equation}
\hspace{-1.7mm}
\begin{array}{c}
\alpha_{\text{BES-MP}}(\mbf{x}, \mbf{y}_{\mcl{D}}) \triangleq |\mcl{F}_\star|^{-1} \sum_{f_\star \in \mcl{F}_\star} I(y_{\mbf{x}}; \gamma_{\mbf{x}}^\star| \mbf{y}_{\mcl{D}}, f_\star)\ .
\end{array}
\hspace{-1.2mm}
\label{eq:besmp}
\end{equation}
At first glance, it may not seem straightforward to justify  averaging BES over $\mcl{F}_\star$ in~\eqref{eq:besmp}.
To do so, we have proven in \citesupp{app:besmpasmi} that the average of BES over $\mcl{F}_\star$~\eqref{eq:besmp} is in fact the mutual information between $y_{\mbf{x}}$ and the jointly distributed random variables $(\gamma_{\mbf{x}}^\star, f_\star)$:\footnote{An alternative acquisition function would be the mutual information $I(y_{\mbf{x}}; \gamma_{\mbf{x}}^\star|\mbf{y}_{\mcl{D}})$ where $f_\star$ is marginalized out.
But, its empirical performance does not differ much from that of~\eqref{eq:besmp}. So, we focus on~\eqref{eq:besmp} which can be seamlessly unified with BES for LSE.}
\begin{equation}
\alpha_{\text{BES-MP}}(\mbf{x}, \mbf{y}_{\mcl{D}}) = I(y_{\mbf{x}}; (\gamma_{\mbf{x}}^\star, f_\star)| \mbf{y}_{\mcl{D}})
\label{eq:besmp2}
\end{equation}
where we overload the notation $f_\star$ to denote a discrete uniform random variable on the support $\mcl{F}_\star$ whose distribution approximates that of the unknown maximum value of $f$.

In the rest of this section, we will investigate the connections between BES-MP and existing acquisition functions: UCB \cite{srinivas10ucb} and MES \cite{wang17mes}.
Our result below reveals that UCB can, interesting, be derived from BES-MP by choosing a deterministic estimate of the maximum value of $f$, as proven in \citesupp{app:besmpasucb}:
\begin{theorem}[Connection to UCB]
\label{theo:ucb}
Define acquisition function of UCB as $\alpha_{\text{\emph{UCB}}}(\mbf{x}, \mbf{y}_{\mcl{D}}) \triangleq \mu_{\mbf{x}} + \beta \sigma_{\mbf{x}}$ ($\beta > 0$) and $\mbf{x}_{\text{\emph{UCB}}} \triangleq \argmax_{\mbf{x} \in \mcl{X}} \alpha_{\text{\emph{UCB}}}(\mbf{x}, \mbf{y}_{\mcl{D}})$. 
If observation $y_{\mbf{x}}$ is noiseless and the estimate of maximum value of $f$ is chosen deterministically: $f_\star = \alpha_{\text{\emph{UCB}}}(\mbf{x}_{\text{\emph{UCB}}}, \mbf{y}_{\mcl{D}})$, then BES-MP selects the same input queries as that selected by UCB.
\end{theorem}
For noisy observation $y_{\mbf{x}}$, though both BES-MP and MES employ a set $\mcl{F}_\star$ of samples of the maximum value of $f$, BES-MP differs significantly from MES in both its interpretation and model of noisy observation, as explained in the two remarks below:
\begin{remark}[Interpretation as information gain]
\emph{\mbox{BES-MP} \eqref{eq:besmp2} can be interpreted as information gain on both the class label $\gamma_{\mbf{x}}^\star$ 
and the threshold $f_\star\in \mcl{F}_\star$ inducing the superlevel set $\mcl{X}_{f_\star}^+$ (of potential maximizers) from evaluating $f$ at input query $\mbf{x}$ to observe $y_{\mbf{x}}$. In contrast, MES measures the information gain on maximum value from evaluating $f$ at input query $\mbf{x}$ to observe $y_{\mbf{x}}$. 
BES-MP~\eqref{eq:besmp} is also closely related to BES~\eqref{eq:bes}, thus allowing our unifying information-theoretic framework for BO and LSE to be established.} 
\label{rmk:informationgain}
\end{remark}
\begin{figure}
\includegraphics[width=0.45\textwidth]{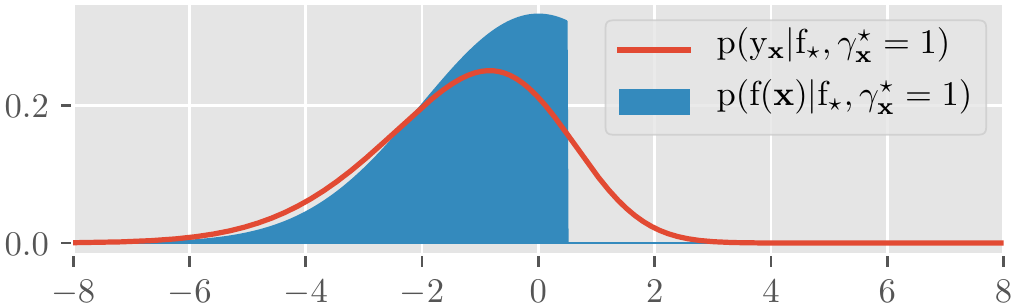}
\caption{Plots of $p(y_{\mbf{x}}| f_\star, \gamma_{\mbf{x}}^\star = 1)$ vs.~a truncated Gaussian distribution of $p(f(\mbf{x})| f_\star, \gamma_{\mbf{x}}^\star = 1)$.}
\label{fig:ntruncnorm}
\end{figure}
\begin{remark}[Model of noisy observation $y_{\mbf{x}}$]
\emph{Another key distinction between BES-MP and MES lies in how they model $p(y_{\mbf{x}}|\mbf{y}_{\mcl{D}}, f_\star, \gamma_{\mbf{x}}^\star = 1)$.
MES assumes that given $f_\star \ge f(\mbf{x})$, the observation $y_{\mbf{x}}$ at an input query $\mbf{x}$ must be at most $f_\star$ \cite{wang17mes}, which leads to an (upper-tail) truncated Gaussian distribution of $p(y_{\mbf{x}}| \mbf{y}_{\mcl{D}}, f_{\star}, \gamma_{\mbf{x}}^\star = 1)$ and its closed-form expression.
However, due to noise $\epsilon_{\mbf{x}}$, $y_{\mbf{x}} = f(\mbf{x}) + \epsilon_{\mbf{x}}$ can be larger than $f_{\star}$ even though $f_{\star} \geq f(\mbf{x})$, as shown in Fig.~\ref{fig:ntruncnorm}.
This issue can be interpreted as MES assuming to observe the noiseless $f(\mbf{x})$ when in fact, only the noisy $y_{\mbf{x}}$ is observed, which implies that MES overestimates the information gain on the maximum value from observing a noisy $y_{\mbf{x}}$. This overestimation is significant when the noise variance $\sigma^2_n$ 
is large relative to the posterior variance $\sigma_{\mbf{x}}^2$~\eqref{eq:gppost} of $f(\mbf{x})$.
Such an issue also plagues the
other MES-based acquisition functions such as those handling multiple objectives \cite{syrine19,suzuki2019multi} or fidelities \cite{takeno2019multi}.
On the other hand, BES-MP models $p(y_{\mbf{x}}|\mbf{y}_{\mcl{D}}, f_\star, \gamma_{\mbf{x}}^\star)$ accurately, which may suggest an improvement to these other MES-based acquisition functions to be considered for future work.}
\label{rmk:besmpmestruncnorm}
\end{remark}
\section{Implicit Level Set Estimation (LSE)}
\label{sec:ilse}
Implicit LSE is about finding the superlevel set w.r.t.~an unknown threshold that differs from the  maximum value of $f$ by a specified \emph{tolerance}.
It is motivated from the estimation of \emph{hotspots} (i.e., superlevel sets) in environmental fields, which are regions of locations (i.e., inputs) whose field measurements (i.e., function values) are of at least a threshold.
Since such measurements may vary throughout the year, it is desirable to define the threshold based on the (unknown) maximum value of the environmental field, which explains the term of \emph{implicit level set}.
For example, farmers are interested to identify the regions of their farms with high (or low) phosphorus level.
Recall that LSE aims to find the superlevel set w.r.t.~a known threshold while BO aims to find the maximizer(s) of the objective function, i.e., the superlevel set w.r.t.~the unknown maximum value.
Therefore, our LSE and BO algorithms cannot be directly applied to solve the implicit LSE problem.

A variant of an implicit LSE problem with a discrete input domain has been introduced in \cite{gotovos2013active} where the threshold is expressed as a percentage of $\max_{\mbf{x} \in \mcl{X}} f(\mbf{x})$.
However, in this paper, we prefer our above definition as it accounts meaningfully for negative function values. 
Note that existing works only consider problems with a discrete input domain \cite{gotovos2013active} while our work here addresses problems with a continuous input domain such as those in our experiments.

Let $\alpha \ge 0$ be the specified tolerance.
The threshold in implicit LSE is then $\max_{\mbf{x} \in \mcl{X}} f(\mbf{x}) - \alpha$ which is not known due to the unknown maximum value: $\max_{\mbf{x} \in \mcl{X}} f(\mbf{x})$.
So, the implicit LSE problem is about finding the superlevel set w.r.t.~$\max_{\mbf{x} \in \mcl{X}} f(\mbf{x}) - \alpha$.
It is a generalized variant of BO and LSE as it reduces to BO when $\alpha = 0$ and to LSE when the maximum value of $f$ is known.

Following the design of BES-MP in Sec.~\ref{sec:besbo}, one may be tempted to solve the implicit LSE problem by averaging BES over the set $\mcl{F}_\alpha \triangleq \{f_\star - \alpha| f_\star \in \mcl{F}_\star\}$ where $\mcl{F}_\star$ is a set of samples of the maximum value of $f$ defined in Sec.~\ref{sec:besbo} previously; $f_\alpha \in \mcl{F}_\alpha$ is then an estimate of the unknown threshold in  implicit LSE. 
Define the superlevel set $\mcl{X}_{f_\alpha}^+ \triangleq \{\mbf{x} \in \mcl{X}| f(\mbf{x}) \ge f_\alpha \}$ w.r.t.~$f_\alpha$.
Let $\gamma_{\mbf{x}}^\alpha$ denote an indicator variable of label $-1$ if $\mbf{x} \in \mcl{X}_{f_\alpha}^+$, and label $1$ otherwise. 
Similar to~\eqref{eq:besmp}, the active learning criterion of BES-MP for implicit LSE can be written as
\begin{equation}
\begin{array}{c}
(1/|\mcl{F}_\alpha|) \sum_{f_{\alpha} \in \mcl{F}_\alpha} I(y_{\mbf{x}}; \gamma_{\mbf{x}}^\alpha | \mbf{y}_{\mcl{D}}, f_\alpha)\ .
\end{array}
\label{eq:ilsebesmp}
\end{equation}
Like~\eqref{eq:besmp2},~\eqref{eq:ilsebesmp} can also be expressed as $I(y_{\mbf{x}}; (\gamma_{\mbf{x}}^\alpha, f_\alpha)| \mbf{y}_{\mcl{D}})$ which can be interpreted as the information gain on both the class label $\gamma_{\mbf{x}}^\alpha$ 
and the threshold $f_\alpha\in \mcl{F}_\alpha$ inducing the superlevel set $\mcl{X}_{f_\alpha}^+$ 
from evaluating $f$ at input query $\mbf{x}$ to observe $y_{\mbf{x}}$.\footnote{We also overload the notation $f_\alpha$ to denote a discrete uniform random variable on the support $\mcl{F}_\alpha$ whose distribution approximates that of the unknown threshold in implicit LSE.} 
We can optimize \eqref{eq:ilsebesmp} in the same manner as~\eqref{eq:besmp}.

\begin{figure}
\hspace{-2.5mm}
\begin{tabular}{cc}
\includegraphics[height=0.235\textwidth]{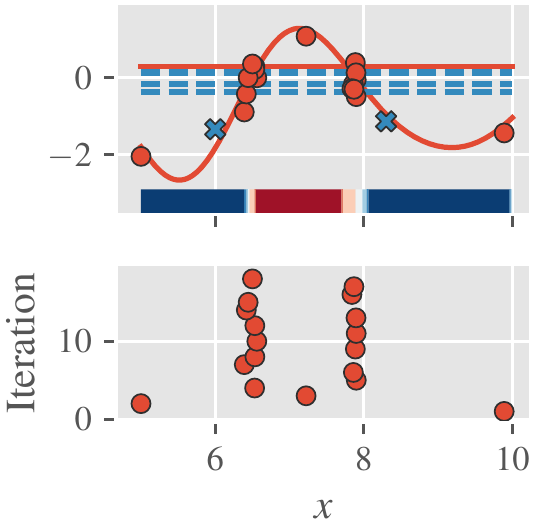}
&
\includegraphics[trim={1cm 0 0 0}, clip, height=0.235\textwidth]{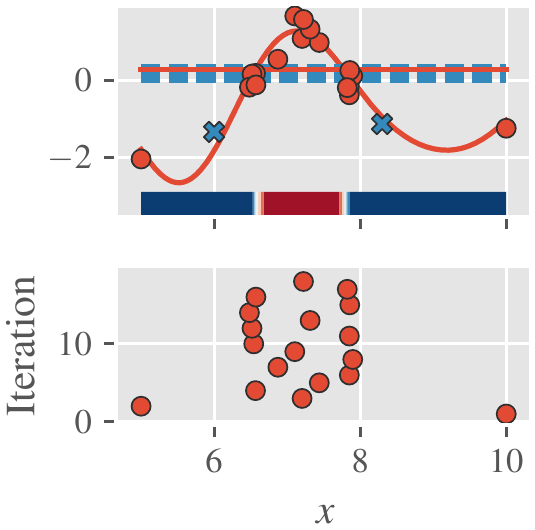}
\\
(a) BES-MP
&
(b) BES$^2$-MP
\end{tabular}
\caption{Input queries of (a) BES-MP and (b) BES$^2$-MP in an implicit LSE problem. 
The notations are the same as those in Fig.~\ref{fig:class} except that the ground truth $f_{\alpha}$ is plotted as a solid red line and the $5$ estimates of $f_{\alpha}$ given $\mcl{D}$ (i.e., after $20$ queries) are plotted as dashed blue lines.}
\label{fig:bes2mp}
\end{figure}

Unfortunately, the above BES-MP only actively estimates the decision boundaries between $\mcl{X}_{f_\alpha}^+$ and $\mcl{X}_{f_\alpha}^-$ for  $f_\alpha \in \mcl{F}_\alpha$. Since these decision boundaries can be far from the maximizer(s) (e.g., when $\alpha$ is large), it is unlikely that BES-MP queries at the maximizer(s), hence yielding poor estimates $f_\star$ of the maximum value. For example,  Fig.~\ref{fig:bes2mp}a shows that BES-MP has only $1$ input query near to the maximizer of $f$. The poor estimates $f_\star$ entail poor estimates $f_\alpha = f_\star - \alpha$ (i.e., dashed blue lines in Fig.~\ref{fig:bes2mp}a) and hence the poor performance of BES-MP in implicit LSE.

To improve the performance of BES-MP in implicit LSE, we consider a generalization of LSE to the \emph{$k$-level set estimation} ($k$-LSE) problem (i.e., with multiple thresholds).
It is an active learning problem that involves actively estimating the $k$ level sets where the threshold of the $i$-th level set is represented by $b_i$. Let $\mbf{b} \triangleq (b_i)_{i=1}^k$ denote a vector of thresholds in ascending order, i.e., $b_i < b_j$ if $i < j$. The $k$-LSE is equivalent to a $(k+1)$-class classification problem that classifies each $\mbf{x} \in \mcl{X}$ into $k+1$ classes. 
Let $\gamma_{\mbf{x}}^k \in \{0,1,\dots,k\}$ denote the class label of an input $\mbf{x}$ such that 
it is of label $0$ if $f(\mbf{x}) \in (-\infty, b_1)$, and label $i$ if $f(\mbf{x}) \in [b_i, b_{i+1})$ and $1 \le i \le k$ where $b_{k+1} \triangleq \infty$.
Similar to the design of BES, we propose an active learning criterion for $k$-LSE called BES$^k$ that measures the information gain on class label $\gamma_{\mbf{x}}^k$ from evaluating $f$ at input query $\mbf{x}$ to observe $y_{\mbf{x}}$:
$$
\alpha_{\text{BES}^k}(\mbf{x}, \mbf{y}_{\mcl{D}}) \triangleq I(y_{\mbf{x}}; \gamma_{\mbf{x}}^k| \mbf{y}_{\mcl{D}}, \mbf{b})
$$
which can be expressed in a form that can be optimized via stochastic gradient ascent (\citesupp{app:besk}).

Implicit LSE can be viewed as a $k$-LSE problem such that the vector of thresholds is unknown (due to the unknown maximum value of $f$).
So, we can exploit our BES$^k$ criterion for $k$-LSE to design an active learning criterion for implicit LSE called BES$^2$-MP (i.e., $k =2$) by averaging BES$^k$ over a set $\mcl{B}$ of estimates $\mbf{b} = (f_\star - \alpha, f_\star)^\top$ for $f_\star \in \mcl{F}_\star$:
$$
\begin{array}{c}
\alpha_{\text{BES}^2\text{-MP}}(\mbf{x}, \mbf{y}_{\mcl{D}})
    \triangleq (1/|\mcl{B}|) \sum_{\mbf{b} \in \mcl{B}} I\left(y_{\mbf{x}}; \gamma_{\mbf{x}}^k | \mbf{y}_{\mcl{D}}, \mbf{b}\right) .
\end{array}
$$
Similar to BES-MP~\eqref{eq:ilsebesmp}, BES$^2$-MP can also be expressed as $I(y_{\mbf{x}}; (\gamma_{\mbf{x}}^k, \mbf{b})| \mbf{y}_{\mcl{D}})$ 
which can be interpreted as the information gain on both the class label $\gamma_{\mbf{x}}^k$ 
and the threshold vector $\mbf{b}$ inducing the $2$ level sets  
from evaluating $f$ at input query $\mbf{x}$ to observe $y_{\mbf{x}}$.\footnote{We also overload the notation $\mbf{b}$ to denote a discrete uniform random variable on the support $\mcl{B}$ whose distribution approximates that of the vector of unknown thresholds.}
Fig.~\ref{fig:bes2mp}b shows that BES$^2$-MP uses several input queries to determine the maximum value of $f$ but BES-MP  (Fig.~\ref{fig:bes2mp}a) does not. As a result, BES$^2$-MP can estimate $f_\star - \alpha$ (i.e., $f_\alpha$) more accurately than BES-MP, which can be observed from Fig.~\ref{fig:bes2mp} by comparing the dashed blue lines representing $f_\alpha$  samples with the solid red line representing the ground truth threshold.
\begin{remark}[A unifying framework]
\emph{We introduce a unifying framework for LSE, BO, and implicit LSE problems by interpreting our proposed active learning criteria or acquisition function
as information gain on the class label and the threshold vector $\mbf{b}$ of length $k$ from evaluating $f$ at input query $\mbf{x}$ to observe $y_{\mbf{x}}$.
By setting $k = 1$, our unifying framework encompasses BES for LSE when the threshold is known 
(Sec.~\ref{sec:beslse}) and BES-MP for BO
when the threshold is unknown  (Sec.~\ref{sec:besbo}).
By setting $k = 2$, our unifying framework encompasses BES$^2$-MP for implicit LSE 
when the threshold vector is unknown.
}
\label{rmk:unify}
\end{remark}
\begin{figure}[!ht]
\centering
\begin{tabular}{@{}cc@{}}
\includegraphics[trim={0.2cm 0.9cm 0.3cm 0}, clip, height=0.19\textwidth]{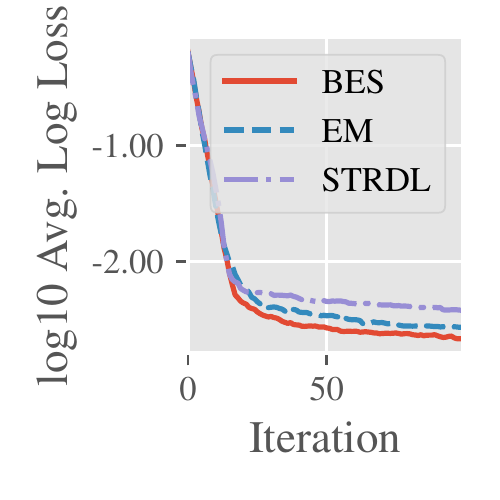}
&
\includegraphics[trim={0.2cm 0.9cm 0.3cm 0}, clip, height=0.19\textwidth]{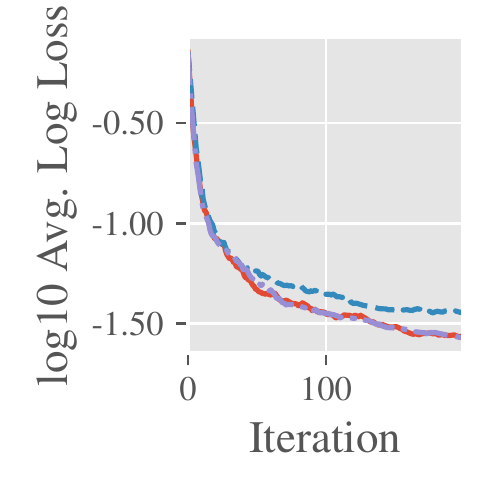}
\\
(a) $\sigma_n^2 = 0.0001$
&
(b) $\sigma_n^2 = 0.09$
\\
\includegraphics[trim={0.2cm 0.9cm 0.3cm 0}, clip, height=0.19\textwidth]{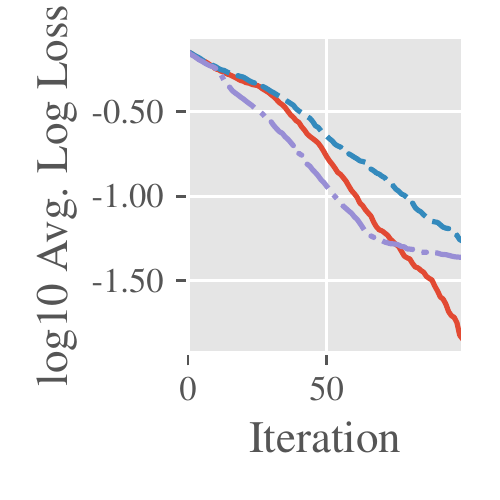}
&
\includegraphics[trim={0.2cm 0.9cm 0.3cm 0}, clip, height=0.19\textwidth]{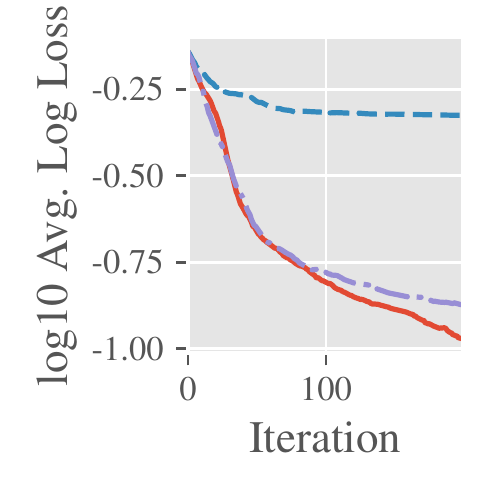}
\\
(c) $\sigma_n^2 = 0.0001$
&
(d) $\sigma_n^2 = 0.09$
\\
\includegraphics[trim={0.2cm 0.9cm 0.3cm 0}, clip, height=0.19\textwidth]{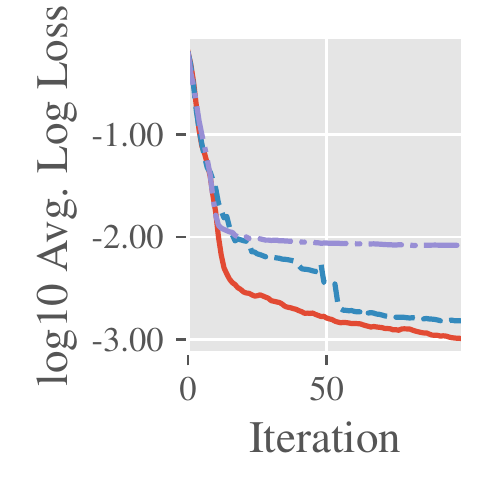}
&
\includegraphics[trim={0.2cm 0.9cm 0.3cm 0}, clip, height=0.19\textwidth]{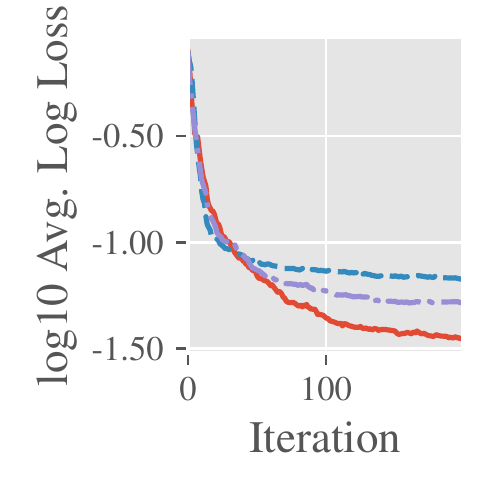}
\\
(e) $\sigma_n^2 = 0.0001$
&
(f) $\sigma_n^2 = 0.09$
\\
\includegraphics[trim={0.2cm 0.9cm 0.2cm 0}, clip, height=0.19\textwidth]{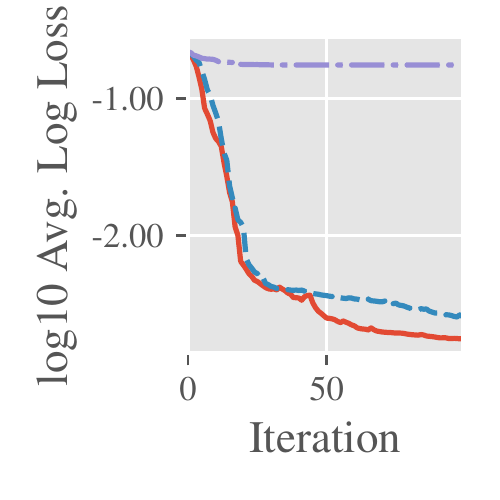}
&
\includegraphics[trim={0.2cm 0.9cm 0.1cm 0}, clip, height=0.19\textwidth]{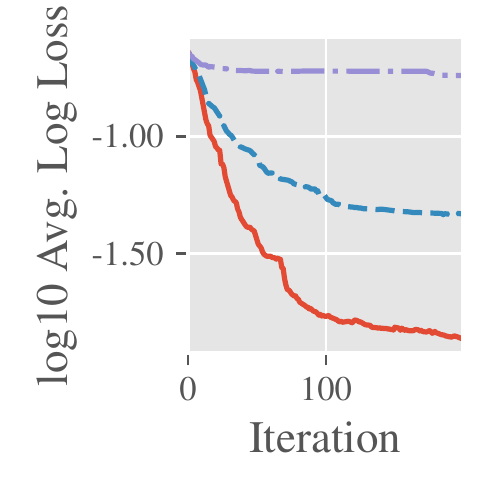}
\\
(g) $\sigma_n^2 = 0.0001$
&
(h) $\sigma_n^2 = 0.09$
\\
\multicolumn{2}{c}{
\includegraphics[trim={0.2cm 0.9cm 0.3cm 0}, clip, height=0.19\textwidth]{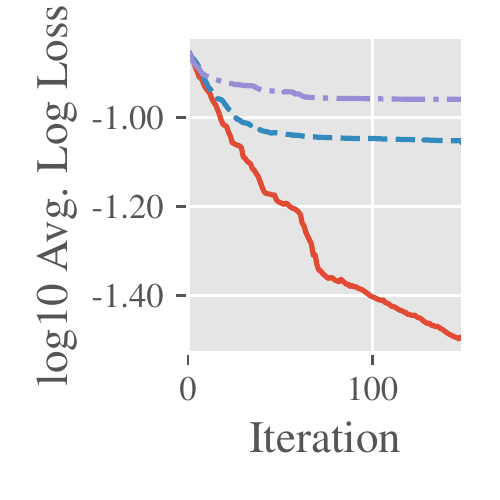}
}
\\
\multicolumn{2}{c}{
(i) $\sigma_n^2 = 0.025$
}
\end{tabular}
\caption{The $\log10$ of the average of the log loss for LSE experiments with synthetic functions: functions drawn from GP with (a-b) $l = 1/3$,  (c-d) $l=0.125$, (e-f) Branin, (g-h) Michaelwicz; and (i) an estimated phosphorus field.}
\label{fig:lsesynthetic}
\end{figure}
\section{Experiments and Discussion}
\label{sec:experiment}
This section empirically evaluates the performance of our proposed LSE (Sec.~\ref{subsec:experimentlse}), BO (Sec.~\ref{subsec:experimentbo}), and implicit LSE  (Sec.~\ref{subsec:experimentilse}) algorithms against that of state-of-the-art methods using synthetic benchmark functions, a real-world dataset, and in hyperparameter tuning of machine learning models. The code is available at \url{https://github.com/qphong/bes-mp}.

\subsection{Level Set Estimation (LSE)}
\label{subsec:experimentlse}
In this subsection, we empirically compare the performance of BES against that of the state-of-the-art EM~\cite{low12} (Remark~\ref{rmk:besaslow})
and \emph{straddle} (STRDL) heuristic \cite{bryan2006active}
in the LSE problem. 
The methods of~\citet{bogunovic16} and~\citet{gotovos2013active} are  demonstrated mainly on problems with a discrete input domain and hence not directly applicable to our experiments with a continuous input domain.
Furthermore, STRDL is empirically shown to achieve comparable performance to these methods.
So, STRDL is chosen as a direct competitor with BES while other methods \cite{bogunovic16,gotovos2013active} are not empirically compared here.
Since LSE is a binary classification problem (see Sec.~\ref{sec:beslse}) in a continuous domain $\mcl{X}$, we use the log loss as the performance metric:
\begin{equation}
\begin{array}{c}
- (1/|\mcl{X}'|) \sum_{\mbf{x}\in\mcl{X}'}  
	\log p(c^{\circ}_{\mbf{x}}\  (f(\mbf{x}) - f_{\circ}) < 0| \mbf{y}_{\mcl{D}})
\end{array}	
\label{eq:lsepmeasure}
\end{equation}
where 
$\mcl{X}'$
is a set of $7000$ uniformly sampled inputs from $\mcl{X}$ and $c^{\circ}_{\mbf{x}}$ is an indicator variable of label $-1$ if $\mbf{x} \in \mcl{X}_{f_\circ}^+$, and label $1$ otherwise.
Each experiment is repeated $30$ times to account for the randomness in the observation and the optimization. Results of the $\log 10$ of the average of the log loss are presented.

As EM assumes noiseless observations (Remark~\ref{rmk:besaslow}),
our experiments are performed with observations of both small ($\sigma_n^2= 0.0001$) and large ($\sigma_n^2=0.09$) noise variances. The GP hyperparameters are learned using \emph{maximum likelihood estimation} (MLE) \cite{rasmussen06}. Regarding the synthetic functions, the function values are normalized and shifted to ensure a zero prior mean.

Results for the synthetic benchmark objective functions\footnote{Details of the synthetic functions are available at \url{https://www.sfu.ca/~ssurjano/optimization.html}.} are shown in Figs.~\ref{fig:lsesynthetic}a to~\ref{fig:lsesynthetic}h.
We can observe that (a) BES outperforms the other active learning criteria for both noise variance values, (b) EM outperforms STRDL when the noise variance is small ($\sigma_n^2 = 0.0001$), as shown in Figs.~\ref{fig:lsesynthetic}a,~\ref{fig:lsesynthetic}e, and~\ref{fig:lsesynthetic}g, and (c) the performance of EM deteriorates when the noise variance is large ($\sigma_n^2 = 0.09$) as it is outperformed by STRDL, as shown in Figs.~\ref{fig:lsesynthetic}b,~\ref{fig:lsesynthetic}d, and~\ref{fig:lsesynthetic}f. The last observation can be explained by the assumption of EM about noiseless observations (Remark~\ref{rmk:besaslow}).

Fig.~\ref{fig:lsesynthetic}i shows the results for an LSE problem on an estimated real-world phosphorus field \cite{webster07}. The noise variance is $\sigma_n^2 = 0.025$ which is learned from the dataset using MLE. It can be observed that BES outperforms EM and STRDL significantly, while EM outperforms STRDL.
The standard deviation (SD) of the log loss is shown in Table~\ref{tbl:lsemstd} in \citesupp{app:mstd}.
\subsection{Bayesian Optimization (BO)}
\label{subsec:experimentbo}
This subsection evaluates the empirical performance of BES-MP against that of the existing acquisition functions: PES, MES, UCB, and EI in optimizing synthetic benchmark functions like Michaelwicz, Hartmann-3d, and Goldstein (the negative values of functions are used), and an estimated environmental field from the phosphorus dataset (see Sec~\ref{subsec:experimentlse}).
The noise variance in the experiments with the synthetic benchmark functions is $0.01$.
The GP hyperparameters are learned using MLE and $|\mcl{F}_\star|$ is set to $5$.

We also use BO to tune the hyperparameters of $2$ machine learning models. Firstly, we train a logistic regression model on the MNIST dataset which consists of $28\times28$ grayscale images of $10$ handwritten digits. The hyperparameters include the L$2$ regularization weight (in $[10^{-6},1]$), the batch size (in $[20,500]$), and the learning rate (in $[10^{-3}, 1]$). So, the input dimension of BO is $3$. The objective function is the validation accuracy on a validation set of $14$K images.
Secondly, we train a CNN on the CIFAR-$10$ dataset which consists of $50$K $32\times32$ color images in $10$ classes. The CNN includes a convolutional layer followed by a dense layer. The hyperparameters include the batch size (in $[32,512]$), the learning rate (in $[10^{-6},10^{-2}]$) and the learning rate decay (in $[10^{-7},10^{-3}]$) of the RMSprop optimization method, the convolutional filter size (in $[128,256]$), and the number of hidden neurons in the dense layer (in $[64,256]$). So, the input dimension of BO is $5$. The objective function is the validation accuracy on a validation set of $10$K images.
We normalize the inputs in these experiments.

Following the work of~\citet{bogunovic16}, the performance metric is the regret of the best input query so far, i.e., $\left( \max_{\mbf{x} \in \mcl{X}} f(\mbf{x}) \right) -  \left( \max_{\mbf{x} \in \mcl{D}} f(\mbf{x}) \right)$.
The regret is averaged over $10$ random runs to account for the randomness in the stochastic optimization and the noisy observation.

\begin{figure}[t]
\centering 
\begin{tabular}{@{}c@{}c@{}}
\includegraphics[trim={0 1cm 0 0.3cm}, clip, height=0.15\textwidth]{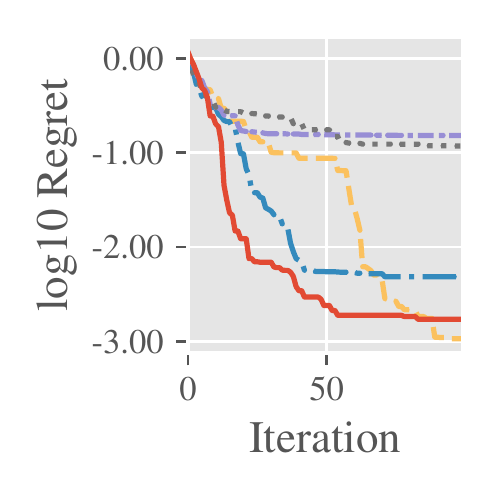}
&
\includegraphics[trim={0.8cm 1cm 0 0.3cm}, clip, height=0.15\textwidth]{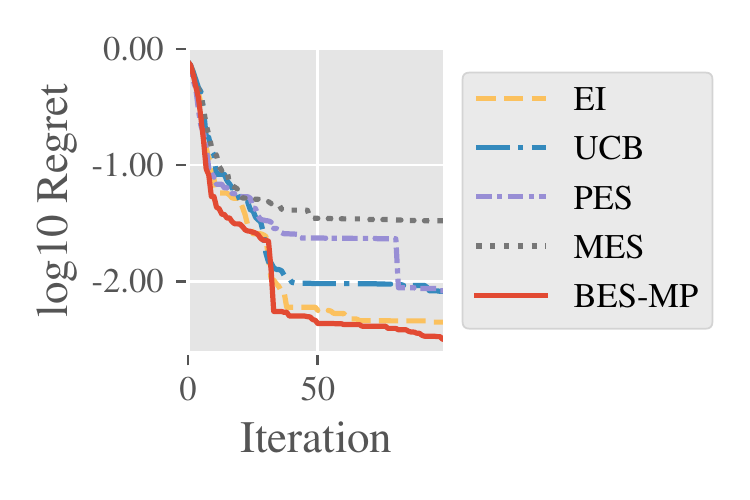}
\\
(a)
&
(b)
\\
\includegraphics[trim={0cm 1cm 0 0.3cm}, clip, height=0.15\textwidth]{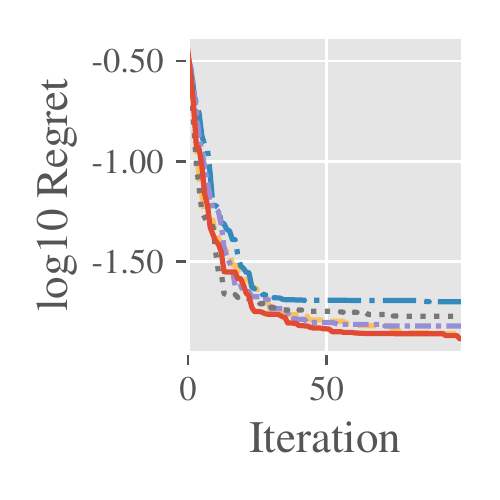}
&
\includegraphics[trim={0.8cm 1cm 0 0.3cm}, clip, height=0.15\textwidth]{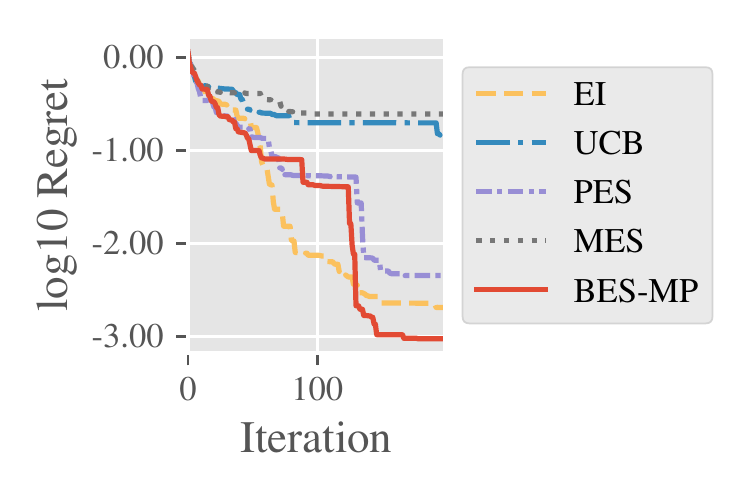}
\\
(c)
&
(d)
\\
\includegraphics[trim={0cm 1cm 0 0.3cm}, clip, height=0.15\textwidth]{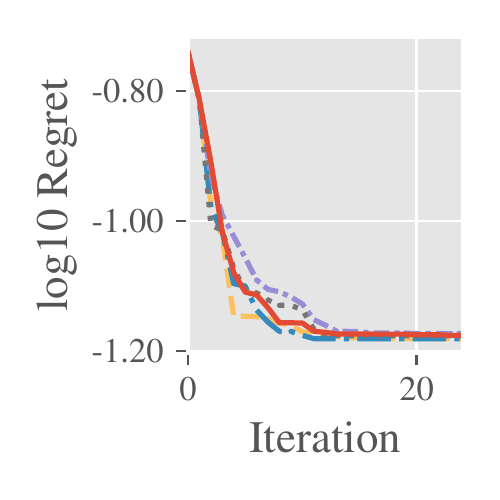}
&
\includegraphics[trim={0.8cm 1cm 0 0.3cm}, clip, height=0.15\textwidth]{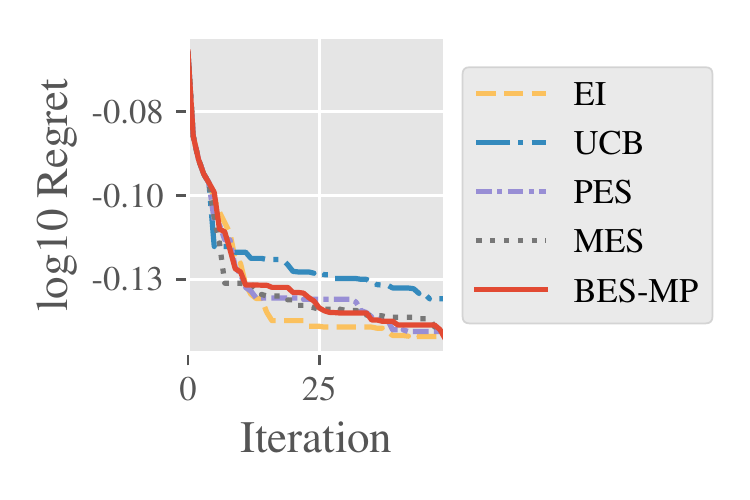}
\\
(e)
&
(f)
\end{tabular}
\caption{BO experiments with synthetic functions:  (a) Michaelwicz, (b) Hartmann-3d, (c) Goldstein; real-world optimization problems: (d) an estimated phosphorus field; in hyperparameter tuning for training (e) a logistic regression model on MNIST, and (f) a CNN on CIFAR-10.}
\label{fig:expalbo}
\end{figure}

Fig.~\ref{fig:expalbo} shows that BES-MP outperforms the other acquisition functions in most of the experiments.
In the other plots, BES-MP demonstrates a comparable performance to that of EI or PES. On the other hand, the performance of MES is not stable: for example, it does not perform well in Figs.~\ref{fig:expalbo}a,~\ref{fig:expalbo}b,~and~\ref{fig:expalbo}d. This can be explained by Remark~\ref{rmk:besmpmestruncnorm}.
The SD of the regret is shown in Table~\ref{tbl:bomstdsr} in \citesupp{app:mstd}.

\subsection{Implicit Level Set Estimation (LSE)}
\label{subsec:experimentilse}

This subsection empirically illustrates the advantage of BES$^2$-MP over BES-MP in implicit LSE problems which include several synthetic benchmark functions and an estimated phosphorus field (see Sec.~\ref{subsec:experimentlse}). The tolerance $\alpha$ is specified as $0.2$. The noise variance $\sigma_n^2$ in the observations of the synthetic functions is $0.0001$.
The GP hyperparameters are optimized using MLE.
The number $|\mcl{F}_\star|$ of maximum value samples is $5$.
Similar to Sec.~\ref{subsec:experimentlse}, the performance metric is the log loss. 
Unlike~\eqref{eq:lsepmeasure}, since the threshold is unknown, it is marginalized out in the log loss expression:
\begin{equation*}
\begin{array}{c}
 -|\mcl{X}'|^{-1} \sum_{\mbf{x} \in \mcl{X}'} \log (  p(c^{\alpha}_{\mbf{x}}\ ( f(\mbf{x}) - f_{\star} + \alpha) < 0 | \mbf{y}_{\mcl{D}}))
 \end{array}
\end{equation*}
where $f_{\star}$ is marginalized: $p(c^{\alpha}_{\mbf{x}}\ ( f(\mbf{x}) - f_{\star} + \alpha) < 0 | \mbf{y}_{\mcl{D}})$ $=|\mcl{F}_{\star}|^{-1} \sum_{f_{\star} \in \mcl{F}_{\star}}  p(c^{\alpha}_{\mbf{x}}\ ( f(\mbf{x}) - f_{\star} + \alpha) < 0 | \mbf{y}_{\mcl{D}}, f_{\star})$;
$c^{\alpha}_{\mbf{x}}$~is an indicator variable of label $-1$ if $\mbf{x} \in \mcl{X}_{f_\alpha}^+$, and label $1$ otherwise.
Each experiment is repeated $30$ times. Results of the $\log 10$ of the average of the log loss are presented.
We also reduce these implicit LSE problems to LSE problems by providing the threshold (i.e., $\max_{\mbf{x} \in \mcl{X}} f(\mbf{x}) - 0.2$) to the active learning criteria for LSE: BES, EM, and STRDL, and plotting their log losses. 
As the threshold is known, these methods serve as baselines that should outperform implicit LSE algorithms, i.e., BES$^2$-MP and BES-MP.

\begin{figure}[t]
\centering
\begin{tabular}{@{}c@{}c@{}}
\multicolumn{2}{@{}c@{}}{
\includegraphics[trim={0.3cm 1cm 0.3cm 0.1cm}, clip, height=0.17\textwidth]{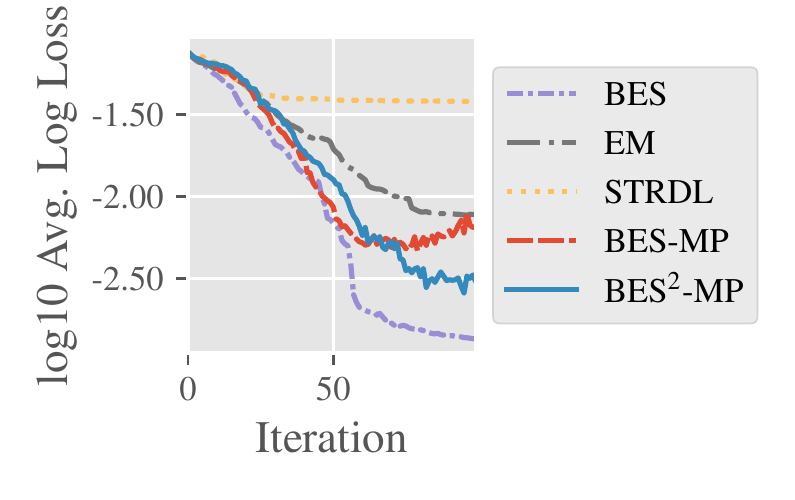}
}
\\
\multicolumn{2}{c}{(a)}
\\
\includegraphics[trim={0cm 1cm 0 0.cm}, clip, height=0.17\textwidth]{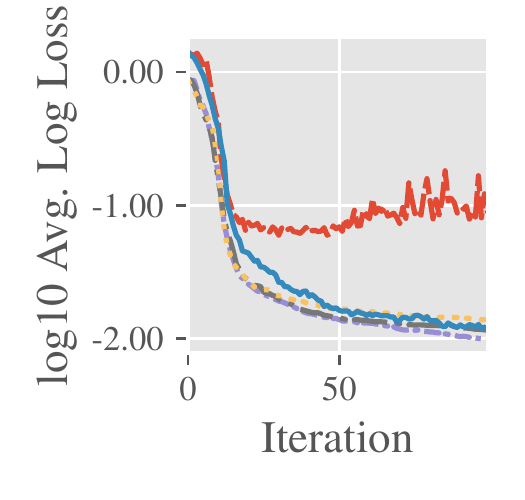}
&
\includegraphics[trim={0cm 1cm 0 0.cm}, clip, height=0.17\textwidth]{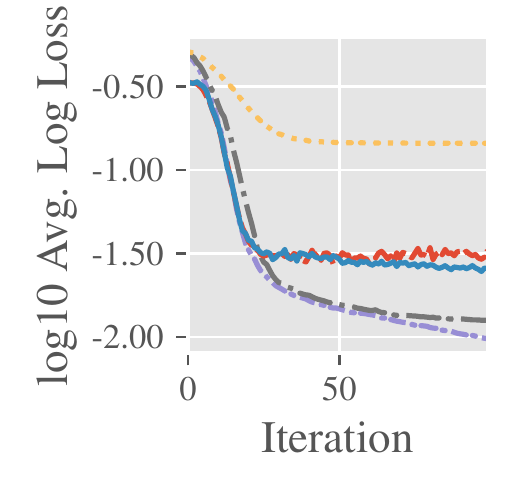}
\\
(b)
&
(c)
\\
\includegraphics[trim={0cm 1cm 0 0.1cm}, clip, height=0.17\textwidth]{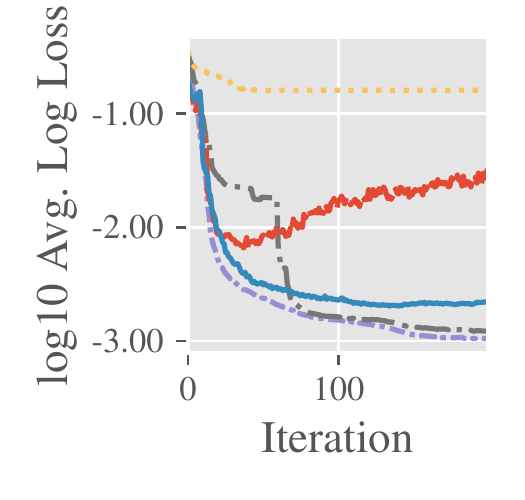}
&
\includegraphics[trim={0cm 1cm 0 0.1cm}, clip, height=0.17\textwidth]{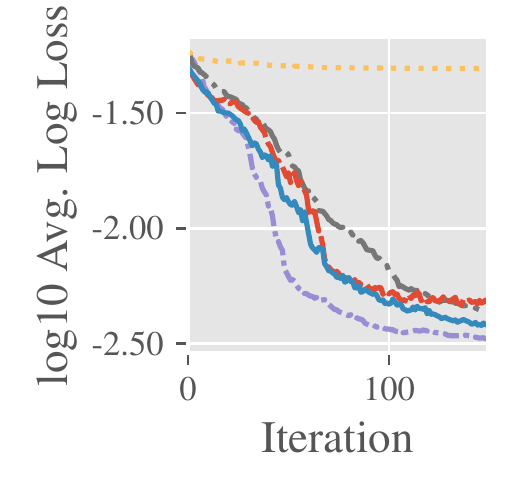}
\\
(d)
&
(e)
\end{tabular}
\caption{Implicit LSE experiments with synthetic functions: (a) a function sampled from GP with $l=0.125$, (b) Branin, (c)  Goldstein, (d) Hartmann-3d; and (e) an estimated phosphorus field. 
}
\label{fig:expregionbo}
\end{figure}

Fig.~\ref{fig:expregionbo} shows that BES$^2$-MP outperforms BES-MP in all experiments, as expected from our discussion in Sec.~\ref{sec:ilse}. Besides, BES-MP does not converge in Figs.~\ref{fig:expregionbo}a~and~\ref{fig:expregionbo}d as BES-MP does not gather observations to learn about the maximum value of $f$ (Sec.~\ref{sec:ilse}).
Regarding the baselines with known thresholds (i.e., active learning criteria for LSE: BES, EM, and STRDL), BES achieves the best performance. However, BES$^2$-MP outperforms EM in Figs.~\ref{fig:expregionbo}a~and~\ref{fig:expregionbo}e likely due to noisy observations.
Surprisingly, even with known thresholds, STRDL is still outperformed by our BES$^2$-MP and BES-MP in several experiments. It is different from the work of \citet{gotovos2013active} where baselines with known thresholds are empirically shown to outperform all methods with unknown thresholds. 
The SD of the log loss is shown in Table~\ref{tbl:mabomstd} in \citesupp{app:mstd}.

\section{Conclusion}
\label{sec:conclusion}
This paper describes an information-theoretic framework for unifying the LSE, BO, and implicit LSE problems.
We propose the first active learning criteria based on mutual information for LSE and implicit LSE problems, which yield the state-of-the-art empirical performance in estimating the level set of synthetic benchmark functions and an environmental field with a continuous input domain.
By exploiting the relationship between LSE and BO, we design an information-theoretic acquisition function and study its connections to UCB and MES. It highlights a critical issue in modeling the noisy observation among the MES-based acquisition functions, which implies their overestimation of the information gain on the maximum value from the noisy observation.
Our proposed acquisition function achieves a competitive performance in comparison with existing  acquisition functions for BO in optimizing synthetic benchmark functions, an environmental field, and in hyperparameter tuning of logistic regression model and CNN.
We will consider generalizing our framework to nonmyopic BO~\cite{dmitrii20a,ling16}, 
batch BO~\cite{daxberger17}, high-dimensional BO~\cite{NghiaAAAI18}, and multi-fidelity BO~\cite{ZhangUAI19} settings.

\section*{Broader Impact}

From our perspective, the societal benefits of the proposed framework outweigh its negative impact.

Our LSE and implicit LSE algorithms can be used for developing methods to monitor/locate hotspots (i.e., regions where environmental field measurements exceed a threshold) in an environmental field (e.g., over lakes and farms), which has potential applications in agriculture, aquaculture, and pollution control.
While some people believe that this development can have a negative impact by reducing the salary of the related jobs, the long-term benefits are more significant.
For example, high-yield and low-cost agriculture can help to sustain the growing population and reduce the food price, which benefits the whole society.

BO is well-known for a wide range of applications such as automated machine learning. With the comparison between our proposed BES-MP and other information-theoretic acquisition functions, other researchers can have a better understanding of BES-MP to employ/enhance it in their own research. Furthermore, our comparison can help engineers to understand and improve existing systems implemented with MES  through the clarification of its drawback, for example, by correcting the approximation in Remark~\ref{rmk:besmpmestruncnorm} if the observation noise is noticeable.

\subsubsection{Acknowledgments.} This research/project is supported by the National Research Foundation, Prime Minister's Office, Singapore under its Campus for Research Excellence and Technological Enterprise (CREATE) program, Singapore-MIT Alliance for Research and Technology (SMART) Future Urban Mobility (FM) IRG. Any opinions, findings, and conclusions or recommendations expressed in this material are those of the author(s) and do not reflect the views of National Research Foundation, Singapore.

    \bibliography{bo}

\begin{thebibliography}{27}
\providecommand{\natexlab}[1]{#1}
\providecommand{\url}[1]{\texttt{#1}}
\providecommand{\urlprefix}{URL }
\expandafter\ifx\csname urlstyle\endcsname\relax
  \providecommand{\doi}[1]{doi:\discretionary{}{}{}#1}\else
  \providecommand{\doi}{doi:\discretionary{}{}{}\begingroup
  \urlstyle{rm}\Url}\fi

\bibitem[{Belakaria, Deshwal, and Doppa(2019)}]{syrine19}
Belakaria, S.; Deshwal, A.; and Doppa, J.~R. 2019.
\newblock Max-value entropy search for multi-objective {Bayesian} optimization.
\newblock In \emph{Proc. {NeurIPS}}, 7825--7835.

\bibitem[{Bogunovic et~al.(2016)Bogunovic, Scarlett, Krause, and
  Cevher}]{bogunovic16}
Bogunovic, I.; Scarlett, J.; Krause, A.; and Cevher, V. 2016.
\newblock Truncated variance reduction: {A} unified approach to {Bayesian}
  optimization and level-set estimation.
\newblock In \emph{Proc. NeurIPS}, 1507--1515.

\bibitem[{Brochu, Cora, and {de Freitas}(2010)}]{brochu10tut}
Brochu, E.; Cora, V.~M.; and {de Freitas}, N. 2010.
\newblock A tutorial on {Bayesian} optimization of expensive cost functions,
  with application to active user modeling and hierarchical reinforcement
  learning.
\newblock {arXiv:1012.2599}.

\bibitem[{Bryan et~al.(2006)Bryan, Nichol, Genovese, Schneider, Miller, and
  Wasserman}]{bryan2006active}
Bryan, B.; Nichol, R.~C.; Genovese, C.~R.; Schneider, J.; Miller, C.~J.; and
  Wasserman, L. 2006.
\newblock Active learning for identifying function threshold boundaries.
\newblock In \emph{Proc. NeurIPS}, 163--170.

\bibitem[{Calandra et~al.(2014)Calandra, Seyfarth, Peters, and
  Deisenroth}]{calandra14}
Calandra, R.; Seyfarth, A.; Peters, J.; and Deisenroth, M.~P. 2014.
\newblock An experimental comparison of {Bayesian} optimization for bipedal
  locomotion.
\newblock In \emph{Proc. {ICRA}}, 1951--1958.

\bibitem[{Daxberger and Low(2017)}]{daxberger17}
Daxberger, E.~A.; and Low, K.~H. 2017.
\newblock Distributed Batch {Gaussian} process optimization.
\newblock In \emph{Proc. {ICML}}, 951--960.

\bibitem[{Galland, R{\'e}fr{\'e}gier, and Germain(2004)}]{galland2004synthetic}
Galland, F.; R{\'e}fr{\'e}gier, P.; and Germain, O. 2004.
\newblock Synthetic aperture radar oil spill segmentation by stochastic
  complexity minimization.
\newblock \emph{IEEE Geoscience and Remote Sensing Letters} 1(4): 295--299.

\bibitem[{Gotovos et~al.(2013)Gotovos, Casati, Hitz, and
  Krause}]{gotovos2013active}
Gotovos, A.; Casati, N.; Hitz, G.; and Krause, A. 2013.
\newblock Active learning for level set estimation.
\newblock In \emph{Proc. IJCAI}, 1344--1350.

\bibitem[{{Hern{\'a}ndez-Lobato}, Hoffman, and Ghahramani(2014)}]{pes}
{Hern{\'a}ndez-Lobato}, J.~M.; Hoffman, M.~W.; and Ghahramani, Z. 2014.
\newblock Predictive entropy search for efficient global optimization of
  black-box functions.
\newblock In \emph{Proc. {NeurIPS}}, 918--926.

\bibitem[{Hoang, Hoang, and Low(2018)}]{NghiaAAAI18}
Hoang, T.~N.; Hoang, Q.~M.; and Low, K.~H. 2018.
\newblock Decentralized high-dimensional {Bayesian} optimization with factor
  graphs.
\newblock In \emph{Proc. {AAAI}}, 3231--3238.

\bibitem[{Hoffman and Ghahramani(2015)}]{hoffman15opes}
Hoffman, M.~W.; and Ghahramani, Z. 2015.
\newblock Output-space predictive entropy search for flexible global
  optimization.
\newblock In \emph{Proc. NeurIPS Workshop on Bayesian Optimization}.

\bibitem[{Kharkovskii, Ling, and Low(2020)}]{dmitrii20a}
Kharkovskii, D.; Ling, C.~K.; and Low, K.~H. 2020.
\newblock Nonmyopic {Gaussian} process optimization with macro-actions.
\newblock In \emph{Proc. AISTATS}, 4593--4604.

\bibitem[{Kingma and Ba(2015)}]{kingma15adam}
Kingma, D.~P.; and Ba, J. 2015.
\newblock Adam: {A} method for stochastic optimization.
\newblock In \emph{Proc. ICLR}.

\bibitem[{Kingma and Welling(2013)}]{kingma2013auto}
Kingma, D.~P.; and Welling, M. 2013.
\newblock Auto-encoding variational {B}ayes.
\newblock {arXiv:1312.6114}.

\bibitem[{Krause and Ong(2011)}]{krause11}
Krause, A.; and Ong, C.~S. 2011.
\newblock Contextual {Gaussian} process bandit optimization.
\newblock In \emph{Proc. NeurIPS}, 2447--2455.

\bibitem[{Ling, Low, and Jaillet(2016)}]{ling16}
Ling, C.~K.; Low, K.~H.; and Jaillet, P. 2016.
\newblock {Gaussian} process planning with {Lipschitz} continuous reward
  functions: Towards unifying {Bayesian} optimization, active learning, and
  beyond.
\newblock In \emph{Proc. {AAAI}}, 1860--1866.

\bibitem[{Low et~al.(2012)Low, Chen, Dolan, Chien, and Thompson}]{low12}
Low, K.~H.; Chen, J.; Dolan, J.~M.; Chien, S.; and Thompson, D.~R. 2012.
\newblock Decentralized active robotic exploration and mapping for
  probabilistic field classification in environmental sensing.
\newblock In \emph{Proc. AAMAS}, 105--112.

\bibitem[{Rahimi and Recht(2008)}]{rahimi08random}
Rahimi, A.; and Recht, B. 2008.
\newblock Random features for large-scale kernel machines.
\newblock In \emph{Proc. {NeurIPS}}, 1177--1184.

\bibitem[{Rasmussen and Williams(2006)}]{rasmussen06}
Rasmussen, C.~E.; and Williams, C. K.~I. 2006.
\newblock \emph{Gaussian Processes for Machine Learning}.
\newblock MIT Press.

\bibitem[{Shahriari et~al.(2015)Shahriari, Swersky, Wang, Adams, and {de
  Freitas}}]{shahriari15}
Shahriari, B.; Swersky, K.; Wang, Z.; Adams, R.; and {de Freitas}, N. 2015.
\newblock Taking the human out of the loop: A review of {Bayesian}
  optimization.
\newblock \emph{Proceedings of the {IEEE}} 104(1): 148--175.

\bibitem[{Snoek, Larochelle, and Adams(2012)}]{snoek12}
Snoek, J.; Larochelle, H.; and Adams, R. 2012.
\newblock Practical {Bayesian} optimization of machine learning algorithms.
\newblock In \emph{Proc. NeurIPS}, 2951--2959.

\bibitem[{Srinivas et~al.(2010)Srinivas, Krause, Kakade, and
  Seeger}]{srinivas10ucb}
Srinivas, N.; Krause, A.; Kakade, S.; and Seeger, M. 2010.
\newblock Gaussian process optimization in the bandit setting: {No} regret and
  experimental design.
\newblock In \emph{Proc. ICML}, 1015--1022.

\bibitem[{Suzuki et~al.(2020)Suzuki, Takeno, Tamura, Shitara, and
  Karasuyama}]{suzuki2019multi}
Suzuki, S.; Takeno, S.; Tamura, T.; Shitara, K.; and Karasuyama, M. 2020.
\newblock Multi-objective Bayesian optimization using {Pareto}-frontier
  entropy.
\newblock In \emph{Proc. {ICML}}.

\bibitem[{Takeno et~al.(2020)Takeno, Fukuoka, Tsukada, Koyama, Shiga, Takeuchi,
  and Karasuyama}]{takeno2019multi}
Takeno, S.; Fukuoka, H.; Tsukada, Y.; Koyama, T.; Shiga, M.; Takeuchi, I.; and
  Karasuyama, M. 2020.
\newblock Multi-fidelity {B}ayesian optimization with max-value entropy search
  and its parallelization.
\newblock In \emph{Proc. {ICML}}.

\bibitem[{Wang and Jegelka(2017)}]{wang17mes}
Wang, Z.; and Jegelka, S. 2017.
\newblock Max-value entropy search for efficient {Bayesian} optimization.
\newblock In \emph{Proc. ICML}, 3627--3635.

\bibitem[{Webster and Oliver(2007)}]{webster07}
Webster, R.; and Oliver, M. 2007.
\newblock \emph{Geostatistics for Environmental Scientists}.
\newblock John Wiley \& Sons, Inc., 2nd edition.

\bibitem[{Zhang, Dai, and Low(2019)}]{ZhangUAI19}
Zhang, Y.; Dai, Z.; and Low, K.~H. 2019.
\newblock Bayesian optimization with binary auxiliary information.
\newblock In \emph{Proc. UAI}.

\end{thebibliography}

\clearpage
\appendix

\section{Derivation of \eqref{eq:bescdf}}
\label{app:evallse}
It is known that $I(y_{\mbf{x}}; \gamma_{\mbf{x}}^\circ | \mbf{y}_{\mcl{D}}, f_{\circ})$ is the \emph{Kullback-Leibler} (KL) divergence between $p(y_{\mbf{x}}, \gamma_{\mbf{x}}^\circ| \mbf{y}_{\mcl{D}}, f_{\circ})$ and $p(y_{\mbf{x}}| \mbf{y}_{\mcl{D}}, f_{\circ})\  p(\gamma_{\mbf{x}}^\circ| \mbf{y}_{\mcl{D}}, f_{\circ})$. So,
\begin{equation}
\hspace{-1.7mm}
\begin{array}{l}
I(y_{\mbf{x}}; \gamma_{\mbf{x}}^\circ | \mbf{y}_{\mcl{D}}, f_{\circ})\vspace{1mm}\\
\displaystyle	= \sum_{\gamma_{\mbf{x}}^\circ} \int p(y_{\mbf{x}}, \gamma_{\mbf{x}}^\circ| \mbf{y}_{\mcl{D}}, f_{\circ})\log \frac{p(y_{\mbf{x}}, \gamma_{\mbf{x}}^\circ| \mbf{y}_{\mcl{D}}, f_{\circ})}{p(y_{\mbf{x}}| \mbf{y}_{\mcl{D}}, f_{\circ})\  p(\gamma_{\mbf{x}}^\circ| \mbf{y}_{\mcl{D}}, f_{\circ})}\ \text{d}y_{\mbf{x}}\vspace{1mm}\\
\displaystyle	= \mbb{E}_{p(y_{\mbf{x}}| \mbf{y}_{\mcl{D}})} \hspace{-1mm}\left[
	\sum_{\gamma_{\mbf{x}}^\circ}  p(\gamma_{\mbf{x}}^\circ| \mbf{y}_{\mcl{D} \cup \{\mbf{x}\}}, f_{\circ}) \log 
		\frac{p(\gamma_{\mbf{x}}^\circ| \mbf{y}_{\mcl{D} \cup \{\mbf{x}\}}, f_{\circ})}{p(\gamma_{\mbf{x}}^\circ| \mbf{y}_{\mcl{D}}, f_{\circ})} \right] .
\end{array}
\label{eq:miygamma}
\end{equation}
Note that 
$$
\begin{array}{l}
\displaystyle p(\gamma_{\mbf{x}}^\circ = -1| \mbf{y}_{\mcl{D}}, f_{\circ}) = 1 - p(\gamma_{\mbf{x}}^\circ = 1| \mbf{y}_{\mcl{D}}, f_{\circ})\vspace{1mm}\\ 
\displaystyle p(\gamma_{\mbf{x}}^\circ = 1| \mbf{y}_{\mcl{D}}, f_{\circ}) = p(f(\mbf{x}) < f_{\circ}| \mbf{y}_{\mcl{D}}, f_{\circ}) = \Psi\hspace{-0.5mm}\left(\frac{f_{\circ} - \mu_{\mbf{x}}}{\sigma_{\mbf{x}}}\right)
\end{array} 
$$
where $\Psi((f_{\circ} - \mu_{\mbf{x}})/ \sigma_{\mbf{x}})$ is the \emph{cumulative density function} (c.d.f.) of the standard Gaussian distribution at \mbox{$(f_{\circ} - \mu_{\mbf{x}}) / \sigma_{\mbf{x}}$}. Then,
\begin{equation}
p(\gamma_{\mbf{x}}^\circ| \mbf{y}_{\mcl{D}}, f_{\circ}) 
	= \Psi\hspace{-0.5mm}\left(\gamma_{\mbf{x}}^\circ\ \frac{f_{\circ} - \mu_{\mbf{x}}}{\sigma_{\mbf{x}}}\right) = \Psi(\gamma_{\mbf{x}}^\circ\  h_{\mbf{x}}(f_{\circ}))
\label{eq:postd}
\end{equation}
where $h_{\mbf{x}}(f_{\circ}) \triangleq (f_{\circ} - \mu_{\mbf{x}}) / \sigma_{\mbf{x}}$.

We can evaluate $p(\gamma_{\mbf{x}}^\circ| \mbf{y}_{\mcl{D}\cup\{\mbf{x}\}}, f_{\circ})$ in the same manner as $p(\gamma_{\mbf{x}}^\circ| \mbf{y}_{\mcl{D}}, f_{\circ})$ by computing the GP posterior belief $p(f(\mbf{x}) | \mbf{y}_{\mcl{D}\cup\{\mbf{x}\}})$ \eqref{eq:gppost} with all the observations $\mbf{y}_{\mcl{D}\cup\{\mbf{x}\}}$, which incurs $\mcl{O}((|\mcl{D}| + 1)^3)$ time.
On the other hand, 
we can compute $p(f(\mbf{x}) | \mbf{y}_{\mcl{D}\cup\{\mbf{x}\}})$ via an incremental update of $p(f(\mbf{x}) | \mbf{y}_{\mcl{D}}) = \mcl{N}(f(\mbf{x})| \mu_{\mbf{x}}, \sigma_{\mbf{x}}^2)$ with the new observation $y_{\mbf{x}}$ as follows:
%
$$
\begin{array}{rl}
p(f(\mbf{x}) | \mbf{y}_{\mcl{D}\cup\{\mbf{x}\}})
	\hspace{-2.4mm} &\displaystyle = \frac{p(f(\mbf{x}) | \mbf{y}_{\mcl{D}})\  p(y_{\mbf{x}}| f(\mbf{x}))}
	{p(y_{\mbf{x}}| \mbf{y}_{\mcl{D}})}\vspace{1mm}\\
	\hspace{-2.4mm} &\displaystyle = \mcl{N} \hspace{-0.5mm}\left( 
		f(\mbf{x}) \Bigg| 
		\frac{\sigma_{\mbf{x}}^2 y_{\mbf{x}} + \sigma_n^2 \mu_{\mbf{x}}}{\sigma_+^2},
		\frac{\sigma_{\mbf{x}}^2 \sigma_n^2}{\sigma_+^2}
	\right)
\end{array}	
$$	
%
where $\sigma_+^2 = \sigma_{\mbf{x}}^2 + \sigma_n^2$ is previously defined in Sec.~\ref{sec:gp}.
As a result,
\begin{equation}
    \begin{array}{l}
\displaystyle p(\gamma_{\mbf{x}}^\circ| \mbf{y}_{\mcl{D}\cup\{\mbf{x}\}}, f_{\circ}) \vspace{1mm}\\
\displaystyle = \Psi\hspace{-0.5mm}\left( \gamma_{\mbf{x}}^\circ \left(f_{\circ} - \frac{\sigma_{\mbf{x}}^2 y_{\mbf{x}} + \sigma_n^2 \mu_{\mbf{x}}}{\sigma_+^2} \right) \Bigg/ \sqrt{\frac{\sigma_{\mbf{x}}^2 \sigma_n^2}{\sigma_+^2}} \right)\vspace{1mm}\\
\displaystyle = \Psi\hspace{-0.5mm}\left( \gamma_{\mbf{x}}^\circ \frac{\sigma_+^2 f_{\circ} - \sigma_n^2 \mu_{\mbf{x}} - \sigma_{\mbf{x}}^2 y_{\mbf{x}}}{\sigma_{\mbf{x}}\sigma_n \sigma_+} \right)\vspace{1mm}\\
\displaystyle = \Psi( \gamma_{\mbf{x}}^\circ\  g_{\mbf{x}}(y_{\mbf{x}}, f_{\circ}))
	\end{array}
\label{eq:postdx}
\end{equation}
where $g_{\mbf{x}}(y_{\mbf{x}}, f_{\circ}) \triangleq \left(\sigma_+^2 f_{\circ} - \sigma_n^2 \mu_{\mbf{x}} - \sigma_{\mbf{x}}^2 y_{\mbf{x}} \right) / \left(\sigma_{\mbf{x}}\sigma_n \sigma_+\right)$.
By plugging~\eqref{eq:postd} and~\eqref{eq:postdx} into \eqref{eq:miygamma}, 
$$
\hspace{-1.7mm}
\begin{array}{l}
I(y_{\mbf{x}}; \gamma_{\mbf{x}}^\circ | \mbf{y}_{\mcl{D}}, f_{\circ}) \vspace{1mm}\\
\displaystyle = \mbb{E}_{p(y_{\mbf{x}}| \mbf{y}_{\mcl{D}})}\hspace{-0.5mm} \left[ 
	\sum_{\gamma_{\mbf{x}}^\circ} \Psi( \gamma_{\mbf{x}}^\circ\  g_{\mbf{x}}(y_{\mbf{x}}, f_{\circ}) )
	 \log \frac{ \Psi( \gamma_{\mbf{x}}^\circ\  g_{\mbf{x}}(y_{\mbf{x}}, f_{\circ}) ) } 
		{\Psi(\gamma_{\mbf{x}}^\circ\  h_{\mbf{x}}(f_{\circ}))}
\right] .
\end{array}
$$
%
\section{Proof of~\eqref{eq:besmp2}}
\label{app:besmpasmi}
In this subsection, we overload the notation $f_\star$ to denote a discrete uniform random variable on the support $\mcl{F}_\star$, i.e., $p(f_\star) = 1/|\mcl{F}_\star|$ for all $f_\star \in \mcl{F}_\star$. We will prove that
$$
I(y_{\mbf{x}}; (\gamma_{\mbf{x}}^\star, f_\star)| \mbf{y}_{\mcl{D}}) = \frac{1}{|\mcl{F}_\star|} \sum_{f_\star \in \mcl{F}_\star} I(y_{\mbf{x}}; \gamma_{\mbf{x}}^\star| \mbf{y}_{\mcl{D}}, f_\star)
$$
where the RHS is the definition of $\alpha_{\text{BES-MP}}(\mbf{x},\mbf{y}_{\mcl{D}})$ in \eqref{eq:besmp} and the LHS is the mutual information in \eqref{eq:besmp2} which allows BES-MP to be interpreted as the information gain on both the class label $\gamma_{\mbf{x}}^\star$ 
and the threshold $f_\star\in \mcl{F}_\star$ inducing the superlevel set $\mcl{X}_{f_\star}^+$ (of potential maximizers)
from evaluating $f$ at input query $\mbf{x}$ to observe $y_{\mbf{x}}$  (Remark~\ref{rmk:informationgain}).

Firstly, we show that $f(\mbf{x})$ and $f_\star$ are conditionally independent if $\gamma_{\mbf{x}}^\star$ is unobserved. We know that
$$
\begin{array}{l}
p(f(\mbf{x})| \mbf{y}_{\mcl{D}}, f_\star)\vspace{1mm}\\
\displaystyle = p(f(\mbf{x}), \gamma_{\mbf{x}}^\star = 1| \mbf{y}_{\mcl{D}}, f_\star) 
+ p(f(\mbf{x}), \gamma_{\mbf{x}}^\star = -1| \mbf{y}_{\mcl{D}}, f_\star)\vspace{1mm}\\
\displaystyle = p(f(\mbf{x})| \mbf{y}_{\mcl{D}}, f_\star, \gamma_{\mbf{x}}^\star=1)\ p(\gamma_{\mbf{x}}^\star = 1| \mbf{y}_{\mcl{D}}, f_\star)\vspace{1mm}\\
\displaystyle \quad +\  p(f(\mbf{x})| \mbf{y}_{\mcl{D}}, f_\star, \gamma_{\mbf{x}}^\star=-1)\ p(\gamma_{\mbf{x}}^\star = -1| \mbf{y}_{\mcl{D}}, f_\star)\ .
\end{array}
$$
We observe that $p(f(\mbf{x})| \mbf{y}_{\mcl{D}}, f_\star, \gamma_{\mbf{x}}^\star=1)$ is a truncated Gaussian probability density function on the support $(-\infty, f_\star)$:
$$
p(f(\mbf{x})| \mbf{y}_{\mcl{D}}, f_\star, \gamma_{\mbf{x}}^\star=1) = \frac{\mbb{I}_{f(\mbf{x}) < f_\star}\  p(f(\mbf{x})| \mbf{y}_{\mcl{D}})}{ p(\gamma_{\mbf{x}}^\star = 1| \mbf{y}_{\mcl{D}}, f_\star)}\ .
$$
Similarly,  
$$
p(f(\mbf{x})| \mbf{y}_{\mcl{D}}, f_\star, \gamma_{\mbf{x}}^\star=-1) = \frac{\mbb{I}_{f(\mbf{x}) \ge f_\star}\  p(f(\mbf{x})| \mbf{y}_{\mcl{D}})}{p(\gamma_{\mbf{x}}^\star = -1| \mbf{y}_{\mcl{D}}, f_\star)}\ .
$$ 
Therefore,
$$
p(f(\mbf{x})| \mbf{y}_{\mcl{D}}, f_\star)
= p(f(\mbf{x})| \mbf{y}_{\mcl{D}})
$$
which implies that $f(\mbf{x})$ and $f_\star$ are conditionally independent if $\gamma_{\mbf{x}}^\star$ is unobserved. Consequently, $y_{\mbf{x}}$ and $f_\star$ are conditionally independent if $\gamma_{\mbf{x}}^\star$ is unobserved: 
$$
\begin{array}{l}
\displaystyle p(y_{\mbf{x}}|\mbf{y}_{\mcl{D}}, f_\star)
	= \int p(y_{\mbf{x}}|f(\mbf{x}))\  p(f(\mbf{x})| \mbf{y}_{\mcl{D}}, f_\star)\ \text{d}y_{\mbf{x}}\vspace{1mm}\\
\displaystyle = \int p(y_{\mbf{x}}|f(\mbf{x}))\  p(f(\mbf{x})| \mbf{y}_{\mcl{D}})\ \text{d}y_{\mbf{x}}
	= p(y_{\mbf{x}}| \mbf{y}_{\mcl{D}})\ .
\end{array}	
$$
It follows that we can express the prior entropy as follows:
$$
\begin{array}{l}
H(p(y_{\mbf{x}}| \mbf{y}_{\mcl{D}}))\\
	\displaystyle = - \int p(y_{\mbf{x}}| \mbf{y}_{\mcl{D}}) \log p(y_{\mbf{x}}| \mbf{y}_{\mcl{D}})\ \text{d}y_{\mbf{x}}\vspace{1mm}\\
	\displaystyle = - \sum_{f_{\star} \in \mcl{F}_{\star}} p(f_{\star}) \int p(y_{\mbf{x}}| \mbf{y}_{\mcl{D}}, f_{\star}) \log p(y_{\mbf{x}}| \mbf{y}_{\mcl{D}}, f_{\star})\ \text{d}y_{\mbf{x}}\vspace{1mm}\\
	\displaystyle = \mbb{E}_{p(f_\star)} \vspace{0.5mm}\left[
	H(p(y_{\mbf{x}}| \mbf{y}_{\mcl{D}}, f_{\star})) \right] .
\end{array}	
$$
Therefore,
$$
\hspace{-1.7mm}
\begin{array}{l} I(y_{\mbf{x}}; (\gamma_{\mbf{x}}^\star, f_{\star})| \mbf{y}_{\mcl{D}})\vspace{1mm}\\
\displaystyle = H(p(y_{\mbf{x}}| \mbf{y}_{\mcl{D}})) - \mbb{E}_{p(\gamma_{\mbf{x}}^\star, f_{\star}|\mbf{y}_{\mcl{D}})} \hspace{-0.5mm}\left[ 
			H(p(y_{\mbf{x}}| \mbf{y}_{\mcl{D}}, \gamma_{\mbf{x}}^\star, f_{\star} ))		
		\right]\vspace{1mm}\\
\displaystyle = \mbb{E}_{p(f_{\star})}\hspace{-0.5mm} \left[ 
	H(p(y_{\mbf{x}}| \mbf{y}_{\mcl{D}}, f_{\star})) \right] - \mbb{E}_{p(\gamma_{\mbf{x}}^\star, f_{\star}|\mbf{y}_{\mcl{D}})}\hspace{-0.5mm} \left[ 
			H(p(y_{\mbf{x}}| \mbf{y}_{\mcl{D}}, \gamma_{\mbf{x}}^\star, f_{\star}))	
		\right]\vspace{1mm}\\
\displaystyle = \mbb{E}_{p(f_{\star})}\hspace{-0.5mm} \left[ 
	H(p(y_{\mbf{x}}| \mbf{y}_{\mcl{D}}, f_{\star})) - \mbb{E}_{p(\gamma_{\mbf{x}}^\star| \mbf{y}_{\mcl{D}}, f_{\star})} \hspace{-0.5mm}\left[ H(p(y_{\mbf{x}}| \mbf{y}_{\mcl{D}}, \gamma_{\mbf{x}}^\star, f_{\star})) \right]
\right]\vspace{1mm}\\
\displaystyle = \mbb{E}_{p(f_{\star})}\hspace{-0.5mm} \left[
	I(y_{\mbf{x}}; \gamma_{\mbf{x}}^\star| \mbf{y}_{\mcl{D}}, f_{\star})
\right] .
\end{array}
$$
Since $f_\star$ follows a discrete uniform distribution on the support $\mcl{F}_\star$, $p(f_\star) = 1 / |\mcl{F}_\star|$. So,
$$
\mbb{E}_{p(f_{\star})}\hspace{-0.5mm} \left[
	I(y_{\mbf{x}}; \gamma_{\mbf{x}}^\star| \mbf{y}_{\mcl{D}}, f_{\star})
\right]
	= \frac{1}{|\mcl{F}_\star|} \sum_{f_\star \in \mcl{F}_\star} I(y_{\mbf{x}}; \gamma_{\mbf{x}}^\star| \mbf{y}_{\mcl{D}}, f_{\star})\ .
$$
\section{Proof of Theorem~\ref{theo:ucb}}
\label{app:besmpasucb}
If the observation is noiseless (i.e., $\sigma_n^2 = 0$) and $f_\star = \alpha_{\text{UCB}}(\mbf{x}_{\text{UCB}},  \mbf{y}_{\mcl{D}})$, then BES-MP reduces to only the prior entropy  $H(p(\gamma_{\mbf{x}}^\star|\mbf{y}_{\mcl{D}}, f_\star))$, as explained in Remark~\ref{rmk:besaslow}.
We will prove that BES-MP selects the same input queries as that selected by UCB:
$$
\begin{array}{c}
\argmax_{\mbf{x} \in \mcl{X}} H(p(\gamma_{\mbf{x}}^\star|\mbf{y}_{\mcl{D}}, f_\star)) = \mbf{x}_{\text{UCB}}\ .
\end{array}
$$
We adapt a proof from that of~\citet{low12} to show that maximizing $H(p(\gamma_{\mbf{x}}^\star|\mbf{y}_{\mcl{D}}, f_\star))$ is equivalent to minimizing $|f_{\star} - \mu_{\mbf{x}}| / \sigma_{\mbf{x}}$ w.r.t.~$\mbf{x}\in \mcl{X}$:
$$
\hspace{-1.7mm}
\begin{array}{l}
\displaystyle\max_{\mbf{x}\in \mcl{X}} H(p(\gamma_{\mbf{x}}^\star|\mbf{y}_{\mcl{D}}, f_\star)) \vspace{1mm}\\
\displaystyle = \max_{\mbf{x}\in \mcl{X}} - \sum_{\gamma_{\mbf{x}}^\star} p(\gamma_{\mbf{x}}^\star| \mbf{y}_{\mcl{D}}, f_{\star}) \log p(\gamma_{\mbf{x}}^\star|\mbf{y}_{\mcl{D}}, f_{\star})\vspace{1mm}\\
\displaystyle = \min_{\mbf{x}\in \mcl{X}} \Big( p(\gamma_{\mbf{x}}^\star = -1| \mbf{y}_{\mcl{D}}, f_{\star}) \log p(\gamma_{\mbf{x}}^\star = -1|\mbf{y}_{\mcl{D}}, f_{\star})\vspace{1mm}\\
\displaystyle \qquad\quad +\  p(\gamma_{\mbf{x}}^\star = 1| \mbf{y}_{\mcl{D}}, f_{\star}) \log p(\gamma_{\mbf{x}}^\star = 1|\mbf{y}_{\mcl{D}}, f_{\star}) \Big)\vspace{1mm}\\
\displaystyle = \min_{\mbf{x}\in \mcl{X}} \Big( p(\gamma_{\mbf{x}}^\star = -1| \mbf{y}_{\mcl{D}}, f_{\star}) \log p(\gamma_{\mbf{x}}^\star = -1|\mbf{y}_{\mcl{D}}, f_{\star})\vspace{1mm}\\
\displaystyle \qquad +\ (1 - p(\gamma_{\mbf{x}}^\star = -1| \mbf{y}_{\mcl{D}}, f_{\star}))\log (1 - p(\gamma_{\mbf{x}}^\star = -1| \mbf{y}_{\mcl{D}}, f_{\star})) \Big)\vspace{1mm}\\
\displaystyle = \min_{\mbf{x}\in \mcl{X}} \left|\frac{1}{2} - p(\gamma_{\mbf{x}}^\star = -1| \mbf{y}_{\mcl{D}}, f_{\star}) \right|\vspace{1mm}\\
\displaystyle = \min_{\mbf{x}\in \mcl{X}} \left|\frac{1}{2} - p(f(\mbf{x}) \ge f_{\star} | \mbf{y}_{\mcl{D}}, f_{\star}) \right|\vspace{1mm}\\
\displaystyle = \min_{\mbf{x}\in \mcl{X}} \left| \text{erf}\hspace{-0.5mm}\left( \frac{f_{\star} - \mu_{\mbf{x}}}{\sigma_{\mbf{x}} \sqrt{2}} \right) \right|\vspace{1mm}\\
\displaystyle = \min_{\mbf{x}\in \mcl{X}} \frac{|f_{\star} - \mu_{\mbf{x}}|}{\sigma_{\mbf{x}}}\ .
\end{array}
$$
That is,  
\begin{equation}
\mathop{\argmax}_{\mbf{x}\in \mcl{X}} H(p(\gamma_{\mbf{x}}^\star|\mbf{y}_{\mcl{D}}, f_\star)) = \mathop{\argmin}_{\mbf{x}\in \mcl{X}} \frac{|f_\star - \mu_{\mbf{x}}|}{\sigma_{\mbf{x}}}\ .
\label{eq:equiemmp}
\end{equation}
Since $\mbf{x}_{\text{UCB}} = \argmax_{\mbf{x} \in \mcl{X}} \alpha_{\text{UCB}}(\mbf{x}, \mbf{y}_{\mcl{D}})$ and $\beta > 0$, 
$$
\begin{array}{rl}
\mu_{\mbf{x}_{\text{UCB}}} + \beta \sigma_{\mbf{x}_{\text{UCB}}}\hspace{-2.4mm} &\ge \mu_{\mbf{x}} + \beta \sigma_{\mbf{x}}\vspace{1mm}\\
\mu_{\mbf{x}_{\text{UCB}}} + \beta \sigma_{\mbf{x}_{\text{UCB}}} - \mu_{\mbf{x}}\hspace{-2.4mm} &\ge \beta \sigma_{\mbf{x}} \ge 0
\end{array}	
$$
for all $\mbf{x} \in \mcl{X}$.
It follows that since $f_{\star} = \alpha_{\text{UCB}}(\mbf{x}_{\text{UCB}}, \mbf{y}_{\mcl{D}}) = \mu_{\mbf{x}_{\text{UCB}}} + \beta \sigma_{\mbf{x}_{\text{UCB}}}$, we can bound  $|f_{\star} - \mu_{\mbf{x}}| / \sigma_{\mbf{x}}$ from below:
$$
\begin{array}{rl}
\displaystyle\frac{|f_{\star} - \mu_{\mbf{x}}|}{\sigma_{\mbf{x}}}\hspace{-2.4mm}
	& \displaystyle =\frac{|\mu_{\mbf{x}_{\text{UCB}}} + \beta \sigma_{\mbf{x}_{\text{UCB}}} - \mu_{\mbf{x}}|}{\sigma_{\mbf{x}}}\vspace{1mm}\\
	& \displaystyle = \frac{\mu_{\mbf{x}_{\text{UCB}}} + \beta \sigma_{\mbf{x}_{\text{UCB}}} - \mu_{\mbf{x}}}{\sigma_{\mbf{x}}}
	\ge \frac{\beta \sigma_{\mbf{x}}}{\sigma_{\mbf{x}}}
	\ge \beta\ .
\end{array}
$$
Furthermore, since $f_{\star} = \alpha_{\text{UCB}}(\mbf{x}_{\text{UCB}}, \mbf{y}_{\mcl{D}}) = \mu_{\mbf{x}_{\text{UCB}}} + \beta \sigma_{\mbf{x}_{\text{UCB}}}$, when $\mbf{x} = \mbf{x}_{\text{UCB}}$, 
$$
\frac{|f_{\star} - \mu_{\mbf{x}_{\text{UCB}}}|}{\sigma_{\mbf{x}_{\text{UCB}}}} = \beta\ .
$$
Therefore,
\begin{equation}
\mathop{\argmin}_{\mbf{x}\in \mcl{X}} \frac{|f_\star - \mu_{\mbf{x}}|}{\sigma_{\mbf{x}}} = \mbf{x}_{\text{UCB}}\ .
\label{eq:minvalue}
\end{equation}
From~\eqref{eq:equiemmp} and~\eqref{eq:minvalue}, we have shown that when $f_{\star} = \alpha_{\text{UCB}}(\mbf{x}_{\text{UCB}}, \mbf{y}_{\mcl{D}})$,  $\argmax_{\mbf{x}\in \mcl{X}} H(p(\gamma_{\mbf{x}}^\star|\mbf{y}_{\mcl{D}}, f_\star)) = \mbf{x}_{\text{UCB}}$.
\section{Alternative Form of BES$^k$}
\label{app:besk}
It is known that $I(y_{\mbf{x}}; \gamma_{\mbf{x}}^k | \mbf{y}_{\mcl{D}}, \mbf{b})$ is the KL divergence between $p(y_{\mbf{x}}, \gamma_{\mbf{x}}^k | \mbf{y}_{\mcl{D}}, \mbf{b})$ and $p(y_{\mbf{x}}| \mbf{y}_{\mcl{D}}, \mbf{b})\  p(\gamma_{\mbf{x}}^k | \mbf{y}_{\mcl{D}}, \mbf{b})$. So, we can obtain a similar expression to~\eqref{eq:miygamma} (Appendix~\ref{app:evallse}):
\begin{equation}
\hspace{-1.7mm}
\begin{array}{l}
I(y_{\mbf{x}}; \gamma_{\mbf{x}}^k | \mbf{y}_{\mcl{D}}, \mbf{b}) \vspace{1mm}\\
\displaystyle = \mbb{E}_{p(y_{\mbf{x}}| \mbf{y}_{\mcl{D}})}\hspace{-0.5mm} \left[\sum_{\gamma_{\mbf{x}}^k}  p(\gamma_{\mbf{x}}^k| \mbf{y}_{\mcl{D} \cup \{\mbf{x}\}}, \mbf{b}) \log 
		\frac{p(\gamma_{\mbf{x}}^k| \mbf{y}_{\mcl{D} \cup \{\mbf{x}\}}, \mbf{b})}{p(\gamma_{\mbf{x}}^k| \mbf{y}_{\mcl{D}}, \mbf{b})} \right] .
\end{array}		
\label{eq:kmiygamma}
\end{equation}
Let $b_0 \triangleq -\infty$ and $b_{k+1} \triangleq \infty$. Then,  $p(\gamma_{\mbf{x}}^k| \mbf{y}_{\mcl{D}}, \mbf{b})$ and $p(\gamma_{\mbf{x}}^k| \mbf{y}_{\mcl{D} \cup \{\mbf{x}\}}, \mbf{b})$ can be expressed as follows:
$$
\begin{array}{l}
\displaystyle p(\gamma_{\mbf{x}}^k| \mbf{y}_{\mcl{D}}, \mbf{b})\vspace{1mm}\\
	\displaystyle = p(b_{\gamma_{\mbf{x}}^k} \le f(\mbf{x}) < b_{\gamma_{\mbf{x}}^k+1}| \mbf{y}_{\mcl{D}}, \mbf{b})\vspace{1mm}\\
	\displaystyle = \Psi\hspace{-0.5mm}\left( \frac{b_{\gamma_{\mbf{x}}^k+1} - \mu_{\mbf{x}}}{\sigma_{\mbf{x}}} \right) - \Psi\hspace{-0.5mm}\left( \frac{b_{\gamma_{\mbf{x}}^k} - \mu_{\mbf{x}}}{\sigma_{\mbf{x}}} \right)\vspace{1mm}\\
	\displaystyle = \Psi( h_{\mbf{x}}(b_{\gamma_{\mbf{x}}^k+1}))
	- \Psi( h_{\mbf{x}}(b_{\gamma_{\mbf{x}}^k}))
\end{array}	
$$
and
$$
p(\gamma_{\mbf{x}}^k| \mbf{y}_{\mcl{D} \cup \{\mbf{x}\}}, \mbf{b})
= \Psi( g_{\mbf{x}}(y_{\mbf{x}}, b_{\gamma_{\mbf{x}}^k+1}))
- \Psi( g_{\mbf{x}}(y_{\mbf{x}}, b_{\gamma_{\mbf{x}}^k}))
$$
where $h_{\mbf{x}}$ and $g_{\mbf{x}}$ are previously defined in the line after \eqref{eq:postd} and \eqref{eq:postdx}, respectively.

We can optimize \eqref{eq:kmiygamma} via  stochastic gradient ascent by reparameterizing the GP posterior belief $p(y_{\mbf{x}}|\mbf{y}_{\mcl{D}})$ to a standard Gaussian distribution \cite{kingma2013auto}.
\onecolumn
\section{Further Experimental Results}
\label{app:mstd}
In this subsection, we present both the mean/average and the standard deviation of the log loss for LSE (Sec.~\ref{subsec:experimentlse}) and implicit LSE (Sec.~\ref{subsec:experimentilse}) experiments and the regret for BO experiments (Sec.~\ref{subsec:experimentbo}) in the last iteration. The results are shown in Tables~\ref{tbl:lsemstd},~\ref{tbl:bomstdsr},~and~\ref{tbl:mabomstd} below:
\begin{table}[!h]
\caption{Mean/average and standard deviation of the log loss for the LSE experiments (Sec.~\ref{subsec:experimentlse}).}
\centering
\begin{tabular}{@{}ccccc@{}}
\toprule
Experiment & $\sigma_n^2$ & BES & EM & STRDL\\
\midrule
\multirow{2}{*}{GP sample ($l=1/3$)} & $0.0001$ & $\mbf{0.0022\pm0.0011}$ & $0.0027\pm0.0015$ & $0.0038\pm0.0012$
\\
    & $0.09$ & $0.0270\pm0.0114$ & $0.0360\pm0.0100$ & $\mbf{0.0265\pm0.0065}$
\\
\\
\multirow{2}{*}{GP sample ($l=0.125$)} & $0.0001$ & $\mbf{0.0136\pm0.0046}$ & $0.0535\pm0.0276$ & $0.0436\pm0.0185$
\\
    & $0.09$ & $\mbf{0.1067\pm0.0192}$ & $0.4722\pm0.1403$ & $0.1339\pm0.0264$
\\
\\
\multirow{2}{*}{Branin} & $0.0001$ & $\mbf{0.0010\pm0.0004}$ & $0.0015\pm0.0006$ & $0.0083\pm0.0140$
\\
    & $0.09$ & $\mbf{0.0354\pm0.0208}$ & $0.0673\pm0.0232$ & $0.0522\pm0.0236$
\\
\\
\multirow{2}{*}{Michaelwicz} & $0.0001$ & $\mbf{0.0017\pm0.0004}$ & $0.0026\pm0.0008$ & $0.1758\pm0.1035$
\\
    & $0.09$ & $\mbf{0.0136\pm0.0041}$ & $0.0467\pm0.0813$ & $0.1815\pm0.0871$
\\
\\
Phosphorus & $0.0251$ & $\mbf{0.0318\pm0.0019}$ & $0.0870\pm0.0430$ & $0.1100\pm0.0438$\\
\bottomrule
\end{tabular}
\label{tbl:lsemstd}
\end{table}
%
\begin{table}[!h]
\caption{Mean/average and standard deviation of the regret for the BO experiments (Sec.~\ref{subsec:experimentbo}).}
\centering
\begin{tabular}{@{}cccccc@{}}
\toprule
Experiment & BES-MP & PES & EI & UCB & MES\\
\midrule
Michaelwicz & $0.0017\pm0.0017$ & $0.1524\pm0.2943$ & $\mbf{0.0011\pm0.0009}$ & $0.0048\pm0.0052$ & $0.1178\pm0.2275$
\\
Hartmann-3d & $\mbf{0.0031\pm0.0017}$ & $0.0083\pm0.0044$ & $0.0044\pm0.0033$ & $0.0082\pm0.0060$ & $0.0332\pm0.0749$
\\
Goldstein & $\mbf{0.0131\pm0.0033}$ & $0.0152\pm0.0037$ & $0.0139\pm0.0029$ & $0.0200\pm0.0118$ & $0.0170\pm0.0042$
\\
Phosphorus & $0.0012\pm0.0013$ & $0.0034\pm0.0029$ & $\mbf{0.0011\pm0.0009}$ & $0.1886\pm0.3068$ & $0.2447\pm0.3182$
\\
MNIST & $0.0667\pm0.0030$ & $0.0669\pm0.0030$ & $0.0659\pm0.0001$ & $\mbf{0.0659\pm0.0000}$ & $0.0660\pm0.0002$
\\
CIFAR-10 & $\mbf{0.376\pm0.022}$ & $0.378\pm0.018$ & $0.377\pm0.010$ & $0.395\pm0.023$ & $0.381\pm0.022$
\\
\bottomrule
\end{tabular}
\label{tbl:bomstdsr}
\end{table}
\begin{table}[!h]
\caption{Mean/average and standard deviation of the log loss for the implicit LSE experiments (Sec.~\ref{subsec:experimentilse}).}
\centering
\begin{tabular}{@{}cccccc@{}}
\toprule
    \multirow{2}{*}{Experiment} & \multicolumn{2}{c}{Unknown $f_\star$ (i.e., implicit LSE)}
    &
    \multicolumn{3}{c}{Known $f_\star$ (i.e., reducing implicit LSE to LSE)}
    \\
\cmidrule(l){2-3} \cmidrule(l){4-6}
    & BES$^2$-MP & BES-MP & BES & EM & STRDL\\
    \midrule
    GP sample & $\mbf{0.0016\pm0.0010}$ & $0.0264\pm0.0251$ & $\mbf{0.0010\pm0.0004}$ & $0.0022\pm0.0018$ & $0.0380\pm0.0224$
    \\
    Branin & $\mbf{0.0125\pm0.0047}$ & $0.1355\pm0.2009$ & $\mbf{0.0100\pm0.0024}$ & $0.0115\pm0.0025$ & $0.0140\pm0.0028$
    \\
    Goldstein & $\mbf{0.0251\pm0.0094}$ & $0.0344\pm0.0308$ & $\mbf{0.0097\pm0.0026}$ & $0.0125\pm0.0028$ & $0.1442\pm0.2235$
    \\
    Hartmann-3d & $\mbf{0.0023\pm0.0006}$ & $0.0331\pm0.0365$ & $\mbf{0.0010\pm0.0003}$ & $0.0012\pm0.0003$ & $0.1598\pm0.2098$
    \\
    Phosphorus & $\mbf{0.0032\pm0.0007}$ & $0.0045\pm0.0018$ & $\mbf{0.0029\pm0.0008}$ & $0.0037\pm0.0018$ & $0.0491\pm0.186$\\
    \bottomrule
\end{tabular}
\label{tbl:mabomstd}
\end{table}

\end{document}